%% file: main.tex
\documentclass{article}

\usepackage[final]{neurips_2023}

\usepackage[utf8]{inputenc} % allow utf-8 input
\usepackage[T1]{fontenc}    % use 8-bit T1 fonts
\usepackage{hyperref}       % hyperlinks
\usepackage{url}            % simple URL typesetting
\usepackage{booktabs}       % professional-quality tables
\usepackage{amsfonts}       % blackboard math symbols
\usepackage{nicefrac}       % compact symbols for 1/2, etc.
\usepackage{microtype}      % microtypography
\usepackage{xcolor}         % colors
\usepackage{graphicx}
\usepackage{subcaption}

\usepackage{custom}
\usepackage{multirow}
\usepackage{floatrow}
\newfloatcommand{capbtabbox}{table}[][\FBwidth]
\usepackage{makecell}
\usepackage{amsmath}
\usepackage{amssymb}
\usepackage{mathtools}
\usepackage{amsthm}
\usepackage{wrapfig}

\usepackage[capitalize,noabbrev]{cleveref}

\usepackage{caption}

\title{Conditional Adapters: Parameter-efficient \\ Transfer Learning
with Fast Inference}

\author{%
 Tao Lei\thanks{Correspondence: \texttt{taole@google.com}}\ \ \ 
 Junwen Bai\ \ \ 
 Siddhartha Brahma\ \ \ 
 Joshua Ainslie\ \ \ 
 Kenton Lee\ \ \ 
 \textbf{Yanqi Zhou}\\
 \textbf{Nan Du} \ \ \ 
 \textbf{Vincent Y. Zhao} \ \ \ 
 \textbf{Yuexin Wu} \ \ \ 
 \textbf{Bo Li}\ \ \ 
 \textbf{Yu Zhang}\ \ \ 
 \textbf{Ming-Wei Chang}\\
 \\
 Google\\
}

\begin{document}

\maketitle

\input{sections/abstract}
\input{sections/introduction}

\input{sections/related}

\input{sections/method}
\input{sections/exp_setup}
\input{sections/results}
\input{sections/conclusion}

\bibliography{coda}
\bibliographystyle{plainnat}

% APPENDIX
\newpage
\appendix

\input{sections/appendix}

\end{document}

%% file: sections/abstract.tex
\begin{abstract}
%\mw{
%1. new ways of getting speed accuracy speed trade off (if title can reflect this, it would be great!)
%
%2. new figure explaining add/remove comparisons with pure adapter

%3. Choose one cherry point in each modality. shows that it is general
%}
We propose \fullname (\name), 
%the first 
a parameter-efficient transfer learning method that also improves {\em inference efficiency}.
\name generalizes beyond standard adapter approaches to enable a new way of balancing speed and accuracy using conditional computation.
Starting with an existing dense pretrained model, \name adds sparse activation together with a small number of new parameters and a light-weight training phase.
Our experiments demonstrate that the \name approach provides an unexpectedly efficient way to transfer knowledge.
Across a variety of language, vision, and speech tasks, \name achieves a 2x to 8x inference speed-up compared to the state-of-the-art Adapter approaches with moderate to no accuracy loss and the same parameter efficiency.
\end{abstract}

% Previous abs
% \begin{abstract}
% We propose \fullname (\name), the first parameter-efficient transfer learning method that also gains {\em inference efficiency.}
% \name generalizes beyond standard adapter approaches to enable a new way to balance the speed-accuracy trade-off using conditional computation, while still freezing all the pretrained parameters and only adding a small number of new parameters.
% \name models dynamically decide if each token requires heavy computation, saving significant amount of inference time.
% Somehow surprisingly, existing densely pretrained models can be converted into sparsely-activated \name models with light-weight pretraining and finetuning. 
% \name demonstrates that conditional computation could be an unexpected efficient way to transfer knowledge.
% %from pretraining to downstream tasks.
% Across natural language processing, vision and speech processing, \name can achieve 2x to 8x inference speed up compared to the state-of-the-art adapter approach~\cite{he2021towards} with moderate-to-none accuracy loss and the same parameter efficiency.  
% \end{abstract}

%% file: sections/introduction.tex
\section{Introduction}
\label{sec:intro}

%\tl{Let's keep "parameter efficiency" and use "computational efficiency" then?}
%\mw{but is it clear that we are talking about testing not training here?}
%\tl{CODA can speed up both training and inference. you mean we should only highlight inference? It sounds good to me but just want to confirm here.}
%\mw{ I am also not sure. but I feel that because coda pretraining, we really did not save anything for training, right?}
%\tl{The step time of CODA is much lower than dense model during training. We don't need to highlight this though.}
%\mw{interesting. let me think about it. Maybe kenton should visit this issue as well}
Large pretrained models have achieved groundbreaking results but the main impediment to deploy them has been the cost of adaptation and inference.
Due to the ever growing size of the pretrained models, for example, finetuning has become increasingly expensive as it requires a separate copy of the full model and updates to all parameters for every downstream task.
%Finetuning has been proven to be an effective approach for transferring knowledge from pretraining to downstream  tasks~\cite{howard2018universal,devlin2019bert}.
%However, due to the ever larger sizes of the pretrained models, this method has become increasingly expensive as it requires a separate copy of the full model and updates to all parameters for each downstream task.
Parameter-efficient transfer learning such as Adapter~\citep{houlsby2019parameter} and Prompt Tuning~\citep{lester2021power} have been proposed to address this issue.  These methods only update a small subset of parameters for each downstream task, allowing the model to retain knowledge and avoid catastrophic forgetting~\citep{vu2022overcoming}. Noticeably, these methods can match the accuracy of a fully finetuned model, while achieving better accuracy on out-of-domain data distributions~\citep{lester2021power, awadalla2022exploring}. 
%\mw{In fact, it is not clear not sure if coda maintains all of the advantages of parameter-efficient transfer learning such as better generalization to other domains}

\begin{figure*}[!t]
\begin{floatrow}
\ffigbox[.6\textwidth]{
    \includegraphics[width=\linewidth]{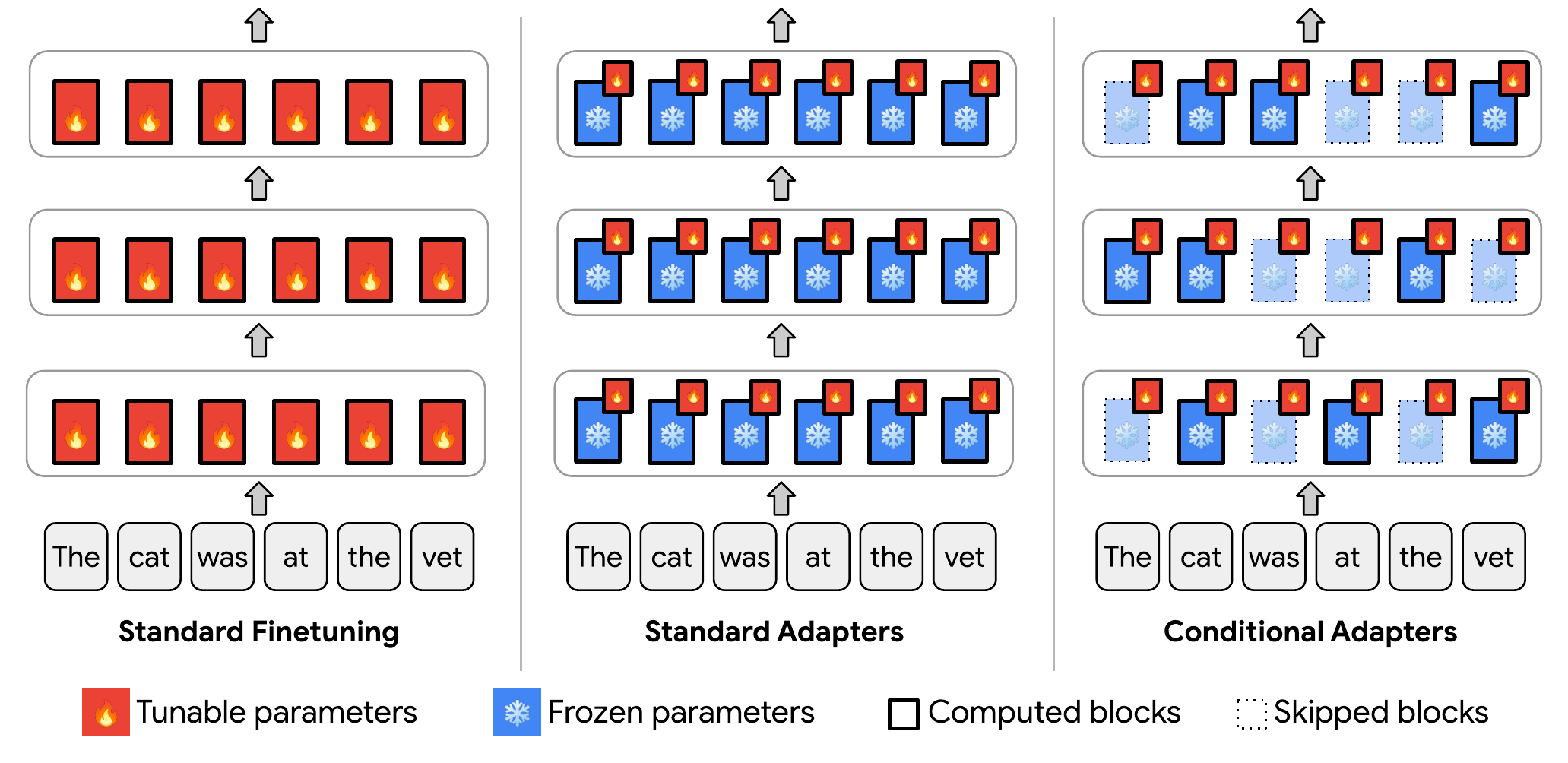}
}{
    \caption{
    Comparison between different ways to use pretrained Transformer models, including (1) standard finetuning (left) where all parameters are tunable and computation is dense, (2) standard adapters (center) where a small set of new tunable parameters are added while the computation remains dense, and (3) \name (right) where the computation is sparsely activated.
    }\label{fig:overview}
}
\capbtabbox[.37\textwidth]{
    \resizebox{.95\linewidth}{!}{
        \centering
        \begin{tabular}{lccc}
        \toprule
        & \multirow{2}{*}{New param} & \multicolumn{2}{c}{MNLI (text)}\\
        \cmidrule{3-4}
        & & Acc$\,\uparrow$ & Speedup \\
        \midrule
        P-Adapter & 0.4\% & 91.5 & 1.0x \\
        \name & 0.4\% & 90.7 & {\bf 3.2x} \\
        \midrule
        & \multirow{2}{*}{New param} & \multicolumn{2}{c}{OCR-VQA (vision)}\\
        \cmidrule{3-4}
        & & EM$\,\uparrow$ & Speedup \\
        \midrule
        P-Adapter & 2.8\% & 67.5 & 1.0x \\
        \name & 2.8\% & 67.6 & {\bf 8.0x} \\
        \midrule
        & \multirow{2}{*}{New param} & \multicolumn{2}{c}{Librispeech (speech)}\\
        \cmidrule{3-4}
        & & WER$\,\downarrow$ & Speedup \\
        \midrule
        P-Adapter & 2.5\% & 1.4/2.7 & 1.0x \\
        \name & 2.5\% & 1.4/2.8 & {\bf 2.2x} \\
        \bottomrule
        \end{tabular}
    }
    %\vspace{0.2in}
}{
    \caption{
    \name significantly reduces the inference time compared to the Parallel Adapter approach~\citep{he2021towards}, while still maintaining parameter efficiency.
    } \label{tab:intro_table}
}
\end{floatrow}
\end{figure*}

Unfortunately, standard parameter-efficient transfer learning methods only bring {\em parameter} efficiency, not {\em inference} efficiency. For example, while only a few small projection matrices are added into the pretrained model in the Adapter approach, all the model inputs (such as tokens) still use all parameters during inference. Therefore, the inference speed is the same (or slightly lower) with respect to the full finetuning method. Moreover, prior studies have shown that these parameter-efficient learning methods are most effective when the size of the pretrained model is large~\citep{lester2021power}, making many advantages of these methods difficult to realize in practice.

In this paper, we propose \fullname (\name),
a parameter-efficient transfer learning method
that offers both {\em parameter} and {\em inference} efficiency. 
\name is a generalization of the adapter approach, built with the following intuition -- we can treat the pretrained model as a universal source of knowledge but only query against it for \emph{necessary inputs}.
Figure~\ref{fig:overview} compares \name with finetuning and standard adapter approaches.
Similar to standard adapter approaches, our model adds and updates a small adapter in each layer, while fixing the pretrained Transformer blocks for downstream adaptation.
Unlike previous approaches, however, \name assumes that many of input token representations (of each layer) are not important for the prediction task and therefore do not require heavy computation.
In such cases, the pretrained Transformer block can be skipped.
Given that many tokens are not processed by the Transformer block, \name runs significantly faster than previous methods.

While conditional activation has clear speed benefits,
\name must learn to select important tokens for heavy computation in order to maintain model accuracy.
To this end, we introduce a soft top-$k$ operation for computing the token selection decision.
This soft top-$k$ operation, which can be seen as a generalization of softmax and a relaxation of hard top-$k$, utilizes entropy-regularized optimization techniques similar to computational optimal transport~\citep{cuturi2013sinkhorn}.
As a result, its output can be computed using fast and differentiable iterations, allowing token selection to be directly optimized for model performance.
%without using additional regularization losses.

%To this end, our model introduces a soft top-$k$ router which can be seen as a generalization of softmax and a relaxation of hard top-$k$.
%Inspired by entropy-regularized optimization such as computational optimal transport~\cite{cuturi2013sinkhorn}, we derive a fast and differentiable algorithm to compute the token selection decision.
%This selection is embedded as an integral part of the model computation and as a result can be directly optimized for downstream performance, without the need of additional regularization loss.

We apply \name on encoder-heavy tasks and evaluate its effectiveness on three different domains -- natural language processing, computer vision and speech processing. Overall, \name achieves 2 to 8 times inference speed-up over standard adapter approach with moderate to no accuracy loss. 
Table~\ref{tab:intro_table} showcases our results by selecting one of the best performing tasks in each domain.
We also conduct comprehensive ablation studies to analyze the effectiveness, efficiency and scalability of \name.
For example, we found that with just a little to no router pretraining, existing dense pretrained models such as T5~\citep{t5} can be efficiently converted into \name models to gain both parameter efficiency and speed advantages.

%% file: sections/related.tex
\section{Related Work}

\paragraph{Parameter-efficient transfer learning methods}
Due to the ever-growing number of parameters in the pretrained Transformer models, various methods have been proposed for transfer learning with minimal parameter updates. 
Prompt tuning~\citep{lester2021power} and prefix tuning~\citep{li2021prefix} introduce new virtual token embeddings that can be finetuned as model parameters. 
Adapter approaches~\citep{houlsby2019parameter, he2021towards} add a small number of new, learnable parameters to each layer while keeping the pretrained parameters fixed. 
Another popular method, Low-Rank Adaptation~\citep[LoRA;][]{hu2021lora}, injects learnable low-rank decomposition matrices into pretrained model parameters.
In addition to requiring less storage cost, parameter-efficient methods have been shown to be more sample-efficient and achieve better out-of-domain generalization than standard finetuning. 
\name is an adapter approach but can be easily combined with other parameter-efficient methods such as LoRA to accelerate their inference.
%\jb{should we re-iterate here inference efficiency is lacking in these methods, and stress on CoDA benefits?}

\paragraph{Conditional computation}
The development of sparsely and conditionally activated models has been a very active research area.
For example, Mixture-of-Experts (MoE) models~\citep{shazeer2017outrageously} and many recent advances~\citep{du2022glam,fedus2021switch} have been proposed to scale up the size of language models without increasing the computation cost.
Many recent works have explored better token routing methods for MoE models, for example using random hashing~\citep{roller2021hash}, balanced assignment~\citep{lewis2021base} and expert-choosing router~\citep{expertchoice2022}.
\name applies conditional computation to both attention and feed-forward blocks of the model, whereas MoE models only focus on sparse activation in the feed-forward blocks.

Similar to our approach, various recent methods have achieved computation efficiency by skipping computation on a subset of input tokens.
However, the selection mechanism can be very different, such as using pooling~\citep{nawrot2022dynamic}, token merging~\citep{bolya2023tome}, token pruning~\citep{rao2021dynamicvit,Yin_2022_CVPR}, learned sigmoid gates~\citep{bapna2020controlling} and early exiting~\citep{schuster2022confident}.
While most of the token merging and pruning methods have been proposed for vision tasks, we show that \name is applicable to multiple domains including text, vision and speech.
In addition, token merging and our token selection method are built with different inductive biases and intuition. Token merging leverages redundancies in visual tokens, while token selection assumes a spike of token relevance. That is, only a few tokens are necessary for the prediction task. Another major difference is that \name dynamically routes and updates token representations in each layer, whereas if a token is pruned (or merged), it will never be re-used by subsequent layers. 
We believe our token routing mechanism is more suited for text and speech applications, such as question answering, where different tokens might play important roles in different layers, or given different input queries.

Finally, \name is closely related to a concurrent work, CoLT5~\citep{ainslie2023colt5}, which also utilizes conditional activation (token selection) for inference efficiency. The focus of CoLT5 and \name are very different. CoLT5 specifically tailors its model architecture for long text (e.g. over 16k tokens), for example, by combining local attention with routed attention. The CoLT5 models are pre-trained from scratch and all parameters are finetuned for downstream tasks. 
In comparison, \name is directly initialized and adapted from an already pretrained dense model, and we optimize its performance on parameter-efficient transfer learning.
The strengths of \name and CoLT5 can be combined for long text applications.
%\name focuses on a different application (i.e. large model adaptation) and introduces a differentiable router to enhance trainability and model performance.

\paragraph{Efficient Transformer models}
Many efficient Transformer variants have been proposed to accelerate model computation.
Examples include creating fast attention variants~\citep{wang2020linformer,Beltagy2020Longformer,guo-etal-2022-longt5,hua2022transformer}, searching network architectures~\citep{press2019improving,so2021searching,su2021vision} and utilizing non-attention neural modules for efficiency~\citep{gulati2020conformer,lei-2021-attention}.
\name utilizes conditional computation as an orthogonal approach for efficiency.
%for additional efficiency.

\paragraph{Model compression}
Apart from building efficient model architectures,
model compression methods such as pruning~\citep{han2015deep_compression,zhu2017prune,wang-etal-2020-structured,xia2022structured} and distillation~\citep{hinton2015distilling,kim-rush-2016-sequence,turc2019well,lin-etal-2020-autoregressive} can be adopted to speed up model inference.
Compared to these methods, \name retains all model parameters of the pretrained large model, and therefore avoids retraining a new model from scratch or knowledge forgetting caused by parameter removal.
In addition,
\name can be seen as a dynamic version of layer pruning because it can activate different Transformer layers for each token, and can be further combined with distillation to reduce the loss of accuracy caused by conditional computation.

%% file: sections/method.tex
\section{Method}
\label{sec:method}

%We describe \name in this section. We start by introducing our notations, followed by an overview of our model architecture and a detailed description of model components.

%\subsection{Preliminaries}
%We use $F()$ to denote a parameterized neural network and the corresponding function defined by the network.
%For instance, a Transformer layer~\citep{vaswani2017attention} consists of an attention sub-layer $F_\textrm{att}()$ followed by a feed forward sub-layer $F_\textrm{ffn}()$.
%Each layer also employs layer normalization~\citep{ba2016layer}, namely $LN_\textrm{att}()$ and $LN_\textrm{ffn}()$, before applying the attention and feed forward functions.

%Let $\mX^l \in \realnumber^{n\times d}$ be the input of the $l$-th Transformer encoder layer, where $n$ is the number of tokens in the input sequence and $d$ is the hidden size of the model.
%The input to the next layer $\mX^{l+1}$ is computed as follows:
%\begin{align}
%    \bar{\mX}^l &= \mX^l + F_\textrm{att}\left(LN_\textrm{att}\left(\mX^l\right)\right) \notag \\
%    \mX^{l+1} &= \bar{\mX}^l + F_\textrm{ffn}\left(LN_\textrm{ffn}\left(\bar{\mX}^l\right)\right)
%\end{align}
%%We provide the definition of functions $F_\textrm{att}()$, $F_\textrm{ffn}()$ and $F_\textrm{norm}()$ in the Appendix for completeness.
%The parameters associated with these functions in each layer are pretrained following the common practices such as BERT~\citep{devlin2019bert} and T5~\citep{t5}.

\subsection{Architecture}
Throughout this and the experiment section, we build \name on top of parallel adapters~\citep{he2021towards}.
However, note that our method can be generalized to other types of adapters such as sequential adapters~\citep{houlsby2019parameter} and LoRA~\citep{hu2021lora}.
We present additional experimental results using LoRA in Appendix~\ref{appendix:lora_results}.
Figure~\ref{fig:overview_2} illustrates our architecture
and shows how \name computes its output by selecting only a small subset of input tokens to query against the pretrained model.
When parallel adapters are used, \name introduces a small number of learnable parameters in the parallel branches, while
the vast majority of model parameters (associated with the pretrained Transformer layers) remain fixed.
In addition, \name only sends $k = \lceil n/r \rceil$ tokens for heavy processing. 
We define $r > 1$ as the reduction factor, a constant (such as 4) to control the computation saving.

Next, we briefly introduce our notations and describe the computation of \name in detail.
We use $F()$ to denote a parameterized neural network and the corresponding function defined by the network.
For instance, a Transformer layer~\citep{vaswani2017attention} consists of an attention sub-layer $F_\textrm{att}()$ followed by a feed forward sub-layer $F_\textrm{ffn}()$.
Each layer also employs layer normalization~\citep{ba2016layer}, namely $LN_\textrm{att}()$ and $LN_\textrm{ffn}()$, before applying the attention and feed forward functions.
We define $\mX\in \realnumber^{n\times d}$ as the input of a Transformer encoder layer, where $n$ is the number of input tokens and $d$ is the hidden size of the model.

\begin{wrapfigure}{r}{1.7in}
\vspace{-0.2in}
\resizebox{\columnwidth}{!}{
    \includegraphics{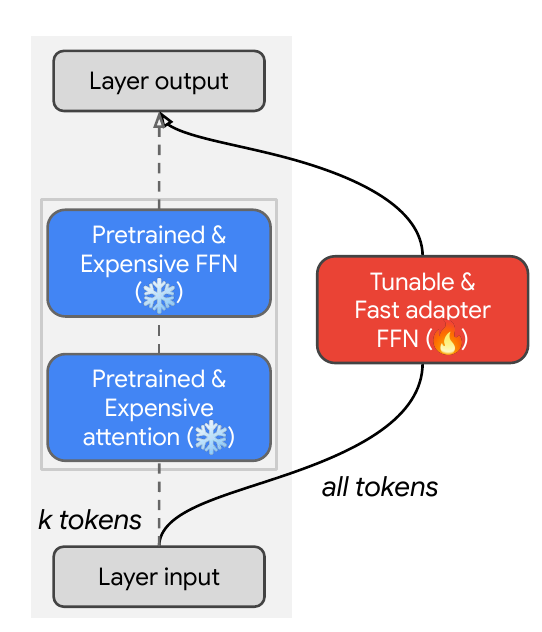}
}
\caption{Illustration of a single \name layer with parallel adapter.
$k$ tokens are selected and processed by the frozen pretrained Transformer layer, and all tokens are processed by the fast adapter layer.}\label{fig:overview_2}
\end{wrapfigure}\vspace{0.0in}

Given layer input
%\footnote{We drop the superscript $l$ when it is clear from the context. All the input and intermediate matrices are defined and computed within the scope of the $l$-th layer.} 
$\mX$, we first apply layer normalization, namely $\mX_\textrm{norm} = LN_\textrm{att}(\mX)$.
The normalized input will be processed by the adapter branch and the conditional Transformer branch. Their outputs are then added and combined as the final output of the layer.

\paragraph{Adapter branch}
Let $F_\textrm{adapter}()$ denote the transformation function of the adapter branch. 
The output is defined as
\begin{align}
    \mZ_\textrm{adapter} = F_\textrm{adapter}(\mX_\textrm{norm})
    %= F_\textrm{adapter}(F_\textrm{norm}(\mX^i))
\end{align}
Similar to the previous approaches, $F_\textrm{adapter}()$ is realized using a feed forward network with a small hidden size such as 64.
As a result, computing $\mZ_\textrm{adapter}$ only incurs a small number of floating point operations and its cost is often negligible compared to the cost of the heavy Transformer branch.
The adapter branch does not conditionally select tokens. In other words, $F_\textrm{adapter}()$ is applied to all input tokens $\mX \in \realnumber^{n\times d}$.

%\mw{can add more structures/visual clues in the next part so reader knows which parts are step 1 2 3. consider giving these steps a name}

\paragraph{Conditional branch}
The computation of the conditional branch takes three steps.
First, each \name layer defines a router function $F_\textrm{router}()$ to select $k$ tokens for the conditional branch.
%where the pretrained Transformer layer is used.
The router function in each layer returns two outputs
\begin{align}
    \vm, \mP = F_\textrm{router}(\mX_\textrm{norm})
    \label{eq:router_outputs}
\end{align}
where $\mP \in \{0, 1\}^{k\times n}$ is a matrix consisting of $k$ one-hot vectors indicating the selection of tokens.
Here $\mP[i, j]=1$ if and only if the $i$-th selected token is the $j$-th input token from $\tilde{\mX}$.
$\vm \in [0, 1]^n$ is a weight mask in which $\vm[j]$ is the selection weight for the $j$-th input token. 
$\vm[j]=0$ if the token is not selected.
We will describe how the router learns the selection in more details later in this section.

After the routing decision is made, the input representations of the selected tokens can be collected using a matrix multiplication,
\begin{align}
\mX_\textrm{routed} &= \mP \mX_\textrm{norm} \ \ \ \ \in \realnumber^{k\times d}
\end{align}
where $k$ rows in $\mX_\textrm{norm}$ are selected to construct the $k$-by-$d$ matrix $\mX_\textrm{routed}$. 
Similar to a standard Transformer layer, the conditional branch applies attention and feed forward transformations to the selected input:
\begin{align}
    \bar{\mZ}_\textrm{routed} &= F_\textrm{att}(\mX_\textrm{routed}) \\
    \mZ_\textrm{routed} &= F_\textrm{ffn}(LN_\textrm{ffn}(\mX_\textrm{routed} + \bar{\mZ}_\textrm{routed}))
\end{align}
where $\bar{\mZ}_\textrm{routed},\mZ_\textrm{routed}\in \realnumber^{k\times d}$ denote the output of the attention network and the feed forward network respectively.
%\footnote{Note we do not apply normalization over $\hat{\mX}$ because it has already been normalized.}.

%Similar to CoLT5~\cite{ainslie2023colt5}, 
We consider two attention variants which differ in how they compute key-value vectors. 
One variant applies a \emph{$k$-to-$k$ attention} using $\mX_\textrm{routed}$ as both the query vectors and key-value vectors.
The other variant applies a \emph{$k$-to-all attention} using the entire input vectors $\mX_\textrm{norm}$ as the attention keys and values.
The $k$-to-all variant runs slower but obtains higher quality close to the full model.
We compare the performance of the two variants in Section~\ref{sec:ablation_results}.

The attention and feed-forward output $\bar{\mZ}_\textrm{routed}$ and $\mZ_\textrm{routed}$ are combined and projected back to the same shape of the original input
\begin{align}
    \mZ_\textrm{cond} &= \mP^\top (\bar{\mZ}_\textrm{routed} + \mZ_\textrm{routed}) \ \ \ \ \in \realnumber^{n\times d}
\end{align}
Finally $\mZ_\textrm{cond}$ merges with the adapter output and the original input of the current layer to produce the output of the layer:
\begin{align}
    \mY &= \mX + \mZ_\textrm{adapter} + \vm \odot \mZ_\textrm{cond}
\end{align}
$\vm\odot \mZ_\textrm{cond}$ is an element-wise multiplication that scales the rows of $\mZ_\textrm{cond}$ using weight $\vm$.
This operation can be seen as a gating operation, where the hidden state $\mZ_\textrm{cond}[i]$ of the $i$-th token is weighted by the token selection score $\vm[i]$ assigned by the router.
This enables gradient propagation from $\vm$ to the router parameters, such that the token selection can be jointly optimized with other model components during training.

\paragraph{Learned router}
An important ingredient of \name is the router function $F_\textrm{router}()$ that is learned to select a subset of tokens for favorable model performance.
Given the token representation $\mX_\textrm{norm}$, our router first computes dot-product score $\vs = \vw\,\mX_\textrm{norm}^\top$,
%\in \realnumber^n$:
%\begin{align}
%    \vs &= \mX_\textrm{norm}\ \vw
%\end{align}
where $\vw \in \realnumber^d$ is a parameter vector associated with the router in this layer. 
The dot-product score $\vs$ is further normalized by a function $f(): \realnumber^n \rightarrow [0, 1]^n$, and clipped to produce the selection score $\vm$:
\begin{align}
    \bm{\lambda} &= f(\vs) \\
    \vm &= \bm{\lambda} \odot \textrm{Top}(\bm{\lambda}, k) \ \ \ \ \in \realnumber^n
\end{align}
Here $\textrm{Top}(\bm{\lambda}, k) \in \{0, 1\}^n$ is an indicator function which returns a binary mask indicating the top-$k$ values in $\bm{\lambda}$.
The one-hot matrix $\mP$ defined in (\ref{eq:router_outputs}) can be created according to $\textrm{Top}(\bm{\lambda}, k)$.
In short, the highest values of $\bm{\lambda}$ will be selected by the router.
%\tl{We don't need the Top($\lambda$,k) clipping technically speaking. We can remove it from the equation above if it makes the section easier to follow.}
%\jb{are there still cases where the highest scores are not selected?}

Function $f()$ must remain differentiable with respect to its input ($\vs$ in this case) such that we can update the router parameters $\vw$ during training.
One possible choice for $f()$ is the sigmoid activation function which normalizes the values in $\vs$ independently.
However, this does not explicitly model the constraint that we need to select $k$ tokens from $n$ available tokens.
Consider a simple case where $k=1$, a natural choice for $f()$ would be the softmax function.
Since softmax provides global normalization over the input scores, a gradient update to increase one of the scores would also decrease the other scores, a desirable effect for learning top-1 selection.

We hypothesize that a \emph{soft top-$k$} operator that generalizes softmax should be used for general $k > 1$.
This is indeed possible by formalizing soft top-$k$ as the following optimization problem:
\begin{align}
f(\vs) \ :=\  \argmax_{\bm{\lambda}}\  \vs^\top \bm{\lambda} + \epsilon H(\bm{\lambda}) \notag \\
\text{s.t.} \quad \bm{1}^\top \bm{\lambda} = k, \ \ \bm{\lambda}[i]\in [0, 1] \ \ \forall i=1,\dots, n
\label{eq:soft_topk}
\end{align}
Here $H(\bm{\lambda}) = \sum_{i=1}^n -\bm{\lambda}[i] \log \bm{\lambda}[i]$ is a generalized entropy function (applied to any positive vector $\bm{\lambda}$ instead of a distribution), and $\epsilon >0$ is a small coefficient. 

This optimization problem is closely related to the softmax and top-$k$ operation.
Specifically, when $\epsilon=0$, it becomes a linear program which returns $\textrm{Top}(\vs, k)$ as the solution.
In addition, when $k=1$, it can be shown that its solution is $\textrm{softmax}(\bm{\vs}/\epsilon)$.
Broadly speaking, (\ref{eq:soft_topk}) will return a soft top-$k$ mask and the smoothness is controlled by $\epsilon$ (and hence $\epsilon$ must be positive to act as a temperature).

Problem (\ref{eq:soft_topk}) does not have a closed-form solution for an arbitrary $\epsilon > 0$ and $k > 1$, but its solution can be obtained using an iterative algorithm. Specifically, let $a \in \realnumber$ and $\vb\in \realnumber^n$ be two auxiliary variables (which can be initialized to zeros). 
The solution takes the form $\bm{\lambda} = \exp(\frac{\vs + \vb + a}{\epsilon})$. 
The values of $a$ and $\vb$ can be obtained using the following iterative updates:
\begin{align}
    a' &= \epsilon \ln(k) -\epsilon \ln\left(\sum_{i=1}^n \exp\left(\frac{\vs[i] + \vb[i]}{\epsilon}\right)\right) \notag \\
    \vb' &= \min(-\vs -a', 0)
    \label{eq:iterations}
\end{align}
In practice, we use $T=20$ iterations and the function $f(\vs)$ returns $\exp(\frac{\vs + \vb + a}{\epsilon})$ using $a$ and $\vb$ from the last iteration.
The function $f(\vs)$ remain differentiable with respect to $\vs$ using these iterative updates, so we can train the router jointly with other model parameters.
We provide additional discussion and the derivation of the updates in Appendix~\S\ref{appendix:soft_topk}.

\subsection{Training}
\label{sec:training}

%Figure~\ref{fig:training_fig} illustrates a general training framework for \name.
%First, as we will demonstrate in our experiments, \name can be directly initialized from an existing Transformer model.
%This includes pretrained language models such as T5~\citep{t5}, BERT~\citep{devlin2019bert}, and similar architectures used for other modalities such as speech~\citep{pmlr-v162-chiu22a} and vision~\citep{lee2022pix2struct}.
\name can be directly initialized from an existing Transformer model.
Given a pretrained model such as T5~\citep{t5}, the Transformer layers are directly re-used and copied in the conditional branches of \name, and only the adapter and router parameters are randomly initialized.
Because pretraining a large dense model can be expensive, our method reduces the overall training cost.

%\begin{figure}
%    \centering
%    \includegraphics[width=.5\columnwidth]{tables_figures/training.pdf}
%    %\vspace{-0.1in}
%    \caption{\name training procedures: \name can be (1) initialized from existing pretrained, dense model by re-using the Transformer layers, (2) optimized using pretraining objective such as language modeling, and then (3) finetuned on downstream tasks. 
%    }
%    \label{fig:training_fig}
%\end{figure}

The routers and neural network components in \name must co-operate and be optimized for accurate model predictions.
When the available finetuning data is limited, a random initialization for the router (and adapter) parameters can be sub-optimal.
We demonstrate that \name can be further pretrained using the same pretraining objective as the dense model, in order to enhance downstream performance.
Importantly, \name requires significantly fewer training steps during pretraining, since most of its parameters are taken from an already pretrained model.
We show that the cost of \name pretraining can be 10-30x lower than the pretraining of its original dense model. We present this analysis in Section~\ref{sec:ablation_results}.

Finally, we train \name on downstream tasks by only updating the adapter, router and layer normalization parameters.
The size of the adapters is small (e.g. 5M parameters), and each router and layer normalization block only introduces $d$ parameters, where $d$ is the model dimension.
As a result, \name remains parameter-efficient similar to previous adapter and prompt-tuning methods.

%% file: sections/exp_setup.tex
\begin{table*}[!t]
\centering
\resizebox{5.52in}{!}{
\begin{tabular}{lc@{~~~~}cccccccccccc}
\toprule
%\multirow{2}{*}{Model} & \multirow{2}{*}{Reduction $r$} 
& & \multicolumn{3}{c}{Base} & & \multicolumn{3}{c}{Large} & & \multicolumn{3}{c}{XL} & \\
%\multirow{2}{*}{Avg $\Delta$} \\
 \cmidrule{3-5} \cmidrule{7-9} \cmidrule{11-13}
Model & Reduction $r\ $ & MNLI & RTE & BoolQ & & MNLI & RTE & BoolQ & & MNLI & RTE & BoolQ & $\Delta$ Avg \\
\midrule
Parallel Adapter  & \multirow{2}{*}{-} & \multirow{2}{*}{87.1} & \multirow{2}{*}{71.5} & \multirow{2}{*}{77.9} & & \multirow{2}{*}{90.3} & \multirow{2}{*}{84.8} & \multirow{2}{*}{85.8} & & \multirow{2}{*}{91.5} & \multirow{2}{*}{89.9} & \multirow{2}{*}{88.4} & \multirow{2}{*}{$\pm$0.0} \\
(w/o conditional computation) \\
\midrule[1pt]
\bf \name, $k$-to-all attention & \multirow{2}{*}{3} & 86.6 & 72.6 & 76.6 & & 90.2 & 85.9 & 85.1 & & 91.4 & 91.3 & 89.4 & $+$0.2 \\
\name, $k$-to-$k$ attention & & 86.3 & 72.2 & 76.2 & & 89.8 & 87.0 & 83.7 & & 91.4 & 89.5 & 88.2 & $-$0.3 \\
\midrule
\bf \name, $k$-to-all attention & \multirow{2}{*}{5} & 86.0 & 70.8 & 76.0 & & 89.7 & 85.2 & 84.3 & & 91.0 & 91.3 & 87.2 & $-$0.6 \\
\name, $k$-to-$k$ attention & & 82.5 & 70.8 & 75.4 & & 88.1 & 87.0 & 81.8 & & 89.9 & 87.7 & 84.8 & $-$2.1 \\
\bottomrule
\end{tabular}
}
\caption{Results of applying \name to T5 v1.1 models. \name achieves significant computation savings while retaining accuracy close to the dense baseline.
We compare \name to a corresponding parallel adapter method that processes all tokens without conditional computation. We report accuracy on the development set on 3 tasks $\times$ 3 model sizes, and set the number of selected tokens $k=\lceil n/r\rceil$. The last column shows the change on average accuracy with respect to the parallel adapter method. We select the $k$-to-all version as our default (shown in bold).}
\label{table:coda_t5}
\end{table*}

\section{Experimental setup}
\label{sec:experimental_setup}

\name is evaluated on three domains including natural language processing (NLP), computer vision and speech processing, and on a range of applications such as classification, question answering, summarization and speech recognition.
The experiments are organized as follows:
%We first demonstrate the effectiveness of \name and conduct analyses on its design choices (\S\ref{sec:ablation_results}).
%The publicly available T5 models are adopted, and our method is tested in different model sizes.
We first demonstrate the effectivenss of \name conduct analyses on its design choices using the publicly available T5 models (\S\ref{sec:ablation_results}).
In our final results (\S\ref{sec:full_results}), we pretrain Transformer models from scratch and extend our evaluation to vision and speech domains.
%by pre-training dense Transformer models and then applying \name (following the procedure described in \S\ref{sec:training}).
%We also expand our language tasks from text classification to generation tasks.

\paragraph{Datasets}
We use the C4 corpus~\citep{t5} for pretraining text models.
For speech models, we use the LibriLight corpus~\citep{librilight} for pretraining.
Our vision Transformer models use the same data and training procedure in Pix2Struct~\citep{lee2022pix2struct}.
%We fine-tune and evaluate our models on publicly available datasets, which we will describe separately in the result sections.
%We select a representative set of fine-tuning tasks in NLP, vision and speech domains.
Our finetuning datasets for text models include the MNLI~\citep{williams-etal-2018-broad}, RTE~\citep{Dagan:2005,haim2006second,giampiccolo2007third,bentivogli2009fifth}, BoolQ~\citep{clark-etal-2019-boolq}, SQuAD~\citep{rajpurkar-etal-2016-squad} and XSum~\citep{Narayan2018DontGM} datasets.
The speech models are evaluated on the speech recognition task using the LibriSpeech dataset~\citep{panayotov2015librispeech}.
Finally, we use the OCR-VQA~\citep{mishra2019ocr}, DocVQA~\citep{mathew2021docvqa}, and Screen2Words~\citep{screen2words} datasets for vision models.

%% file: sections/results.tex
%\section{Results}
%\label{sec:results}

%\subsection{Application to existing text-only models}

\section{Understanding and Analyzing \name}
\label{sec:ablation_results}

\paragraph{Setup}  
We present several analyses to validate the design choices of \name in this section.
We initialize \name using the version 1.1 release 
of T5 checkpoints\footnote{\url{https://github.com/google-research/text-to-text-transfer-transformer/blob/main/released\_checkpoints.md\#t511}}, and perform \name pretraining using the same setting as the T5 models.
During pretraining, we set routing capacity to $k=192$ given input sequence length $n=512$.
We do not tune the value of $k$ for pretraining, but will report the results of using different $k$ values in finetuning.
We perform 100K gradient steps, which is 10\% of the total number of steps used to train the T5 dense models.
The overall computational cost is over 20x less than the full training of dense models, since \name only applies heavy computation on less than half of the tokens.

For simplicity, we evaluate on classification tasks for various ablation studies of \name.
Specifically, we report results on the MNLI, RTE and BoolQ datasets, and 
test three different model sizes including the Base, Large and XL size of T5.
We will extend our evaluation to generation tasks such as question answering in the full result section (\S\ref{sec:full_results}).

%\begin{figure}
%%    \centering
%    \includegraphics[width=2.6in]{tables_figures/pretrain_steps.pdf}
%    \vspace{-0.1in}
%    \caption{Finetuning accuracy (y-axis) as a function of \name pretraining steps (x-axis). We show results using 0, 20K, 50K and 100K steps for \name pretraining.
%    \name requires as few as 20K steps to obtain competitive finetuning accuracy.
%    }
%    \label{fig:pretrain_steps}
%\end{figure}

\begin{table*}[!t]
\centering
\resizebox{5.52in}{!}{
\begin{tabular}{lc@{~~~~}cccccccccccc}
\toprule
%\multirow{2}{*}{Model} & \multirow{2}{*}{Reduction $r$} 
& & \multicolumn{3}{c}{Base} & & \multicolumn{3}{c}{Large} & & \multicolumn{3}{c}{XL} & \\
%\multirow{2}{*}{Avg $\Delta$} \\
 \cmidrule{3-5} \cmidrule{7-9} \cmidrule{11-13}
Model & Reduction $r\ $ & MNLI & RTE & BoolQ & & MNLI & RTE & BoolQ & & MNLI & RTE & BoolQ & $\Delta$ Avg \\
\midrule
\bf Soft top-$k$ & \multirow{3}{*}{3} & 86.3 & 72.2 & 76.2 & & 89.8 & 87.0 & 83.7 & & 91.4 & 89.5 & 88.2 & $\pm$0.0 \\
Sigmoid gate as $f(\vs)$ & & 85.7 & 70.8 & 72.8 & & 89.2 & 82.3 & 81.0 & & 90.6 & 88.1 & 86.2 & $-$2.0 \\
Truncation -- selecting first $k$ & & 81.1 & 70.8 & 72.7 & & 84.9 & 77.3 & 82.3 & & 85.6 & 84.5 & 85.4 & $-$4.4 \\
\midrule
\bf Soft top-$k$ & \multirow{3}{*}{5} & 82.5 & 70.8 & 75.4 & & 88.1 & 87.0 & 81.8 & & 89.9 & 87.7 & 84.8 & $\pm$0.0 \\
Sigmoid gate as $f(\vs)$ & & 82.9 & 71.5 & 72.1 & & 86.7 & 82.3 & 80.1 & & 88.3 & 87.0 & 82.4 & $-$1.6 \\
Truncation -- selecting first $k$ & & 62.2 & 64.6 & 71.1 & & 64.9 & 70.4 & 75.4 & & 66.6 & 76.2 & 81.1 & $-$12.9 \\
\bottomrule
\end{tabular}
}
\vspace{-0.05in}
\caption{Ablation study on routing methods. 
We use \name $k$-to-$k$ variant for a fair comparison with the truncation method. 
Better routing method delivers better accuracy on various tasks and model sizes tested. We use soft top-$k$ as our default method.}
\label{table:routing_ablation}
\end{table*}

\paragraph{Can \name be fast and accurate?}
Table~\ref{table:coda_t5} presents the finetuning results of \name.
As a comparison, we also report the results of Parallel Adapter, which is similar to \name except that it applies the expensive Transformer layers to all input tokens.
This constitutes an upper-bound, and is a strong baseline that has been reported as the best among a range of adapter and prompt tuning methods~\citep{he2021towards}.
As shown in Table~\ref{table:coda_t5}, \name can achieve 3-5x computation reduction ($r=3, 5$)
in the Transformer layers at a cost of less than 1.0 point 
drop on average accuracy.
As expected, our $k$-to-all attention variant achieves consistently better accuracy than the $k$-to-$k$ variant, since it can access the full attention context.
On the other hand, the $k$-to-$k$ attention variant runs faster in practice, which can be beneficial for tasks with very long inputs.
We select the $k$-to-all version in the final result section (\S\ref{sec:full_results}).

\begin{wrapfigure}{r}{2.3in}\vspace{-0.15in}
\resizebox{\columnwidth}{!}{
    \includegraphics{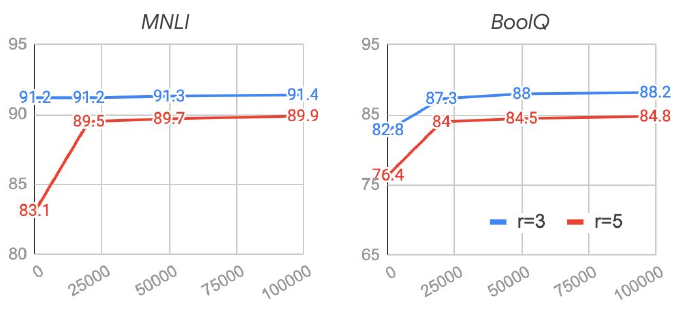}
}
\caption{Finetuning accuracy (y-axis) as a function of \name pretraining steps (x-axis). We show results using 0, 20K, 50K and 100K pretraining steps, and for reduction factor $r=3$ and $r=5$ respectively.
\name requires as few as 20K steps to obtain competitive finetuning accuracy.
}
\label{fig:pretrain_steps}
\end{wrapfigure}
%\vspace{-0.1in}

\paragraph{How many pretraining steps are needed?}
Figure~\ref{fig:pretrain_steps} plots the finetuning accuracy by varying the number of pretraining steps for \name.
Because \name can be initialized using pretrained dense models, it requires as few as 20K steps to obtain competitive finetuning results.
Of course, using more pretraining steps can improve the downstream accuracy.
The fact that \name can be quickly updated without repeating the expensive pretraining will be very beneficial in real-world applications.

\paragraph{Does learned routing matter?}
We analyze the impact of learned routing in Table~\ref{table:routing_ablation} by comparing our soft top-$k$ router with other router implementations.
We implement a variant that replaces soft top-$k$ with the sigmoid activation function, so the selection weight of each token activates on its own (without considering the capacity constraint).
As shown in the table, this variant achieves worse accuracy on almost all tasks and model sizes tested, getting 2.0 point worse on average.
We also implement a ``no-learning'' baseline that simply selects the first $k$ tokens, which is equivalent to truncating the input sequence.\footnote{We always include the question text for BoolQ, to achieve higher accuracy.} 
This baseline performs much worse, resulting in more than 10 point decrease in accuracy for small $k$ (and equivalently large $r$).
This analysis confirms the importance of learning a good routing in order to retain strong model performance.

%\subsection{Application to generation tasks}

%\begin{table*}[!t]
\begin{figure*}[!t]
\begin{floatrow}
\capbtabbox[.76\textwidth]{
    %\small
    %\centering
    \resizebox{\linewidth}{!}{
    \begin{tabular}{lcc@{~~~~}ccccccc}
    \toprule
    \multirow{2}{*}{Model} & Trainable & Reduction & $\,$MNLI$\,$ & $\,$RTE$\,$ & BoolQ & SQuAD & ReCord & $\,$XSum$\,$ & $\Delta$Avg \\
    & Params & $r$ & Acc. & Acc. & Acc. & F1 & F1 & R2 \\
    \midrule
    Parallel Adapter & 10M & - & 91.5 & 91.0 & 88.5 & 94.8 & 91.4 & 21.9 & $\pm$0.0 \\
    \name & 10M & 3 & 91.2 & 90.3 & 87.5 & 94.1 & 89.3 & 20.6 & $-$1.0\\
    \name & 10M & 5 & 90.7 & 89.5 & 87.3 & 93.5 & 87.6 & 20.2 & $-$1.7 \\
    \midrule
    Prefix Tuning~\citep{li2021prefix}$^\dagger$ & 15M (2M) & - & (86.3) & - & - & - & - & 20.5 \\
    Sequential Adapter~\citep{houlsby2019parameter}$^\dagger$ & 10M (2M) & - & (87.2) & - & - & - & - & 20.0 \\
    Parallel Adapter~\citep{he2021towards}$^\dagger$ & 10M & - & - & - & - & - & - & 20.7 \\
%    T5 Fine-tune~\cite{t5} & 2.7B & - & 91.4 & 90.7 & 89.9 & 95.0 & 90.0 & - \\
    \bottomrule
    \end{tabular}
    }
    \vspace{0.25in}
}{
    \caption{Comparison of \name and parallel adapter on 6 language tasks. 
    We report results on the test set of XSum, and on the development set of other tasks. 
    $\dagger$ indicates results taken from \citet{he2021towards}, and referenced results in bracket correspond to using 2M adapter parameters.
    Note that our Parallel Adapter numbers are stronger as our pretrained Transformer backbone uses more parameters than the model used in \citet{he2021towards}. %We report more results in Appendix~\ref{appendix:additional_results}.
    }\label{table:nlp_finetuning_results}
}
\ffigbox[.21\textwidth]{
  \includegraphics[width=.95\linewidth]{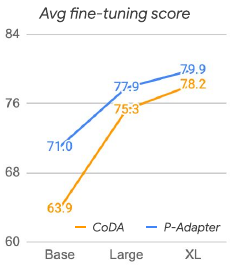}
}{
    \caption{Average finetuning scores of Parallel Adapter and \name at different model sizes. }\label{fig:avg_ft_nlp}
}
\end{floatrow}
\end{figure*}
%\end{table*}

\section{Full Results}
\label{sec:full_results}

\paragraph{Setup}
In this section, we apply our best training recipe to all tasks and application domains.
We first pretrain dense Transformer models, followed by the \name training procedure in \S\ref{sec:training}.
Our speech models are pretrained using a masked language modeling (MLM) objective similar to BERT~\citep{devlin2019bert}, and random quantized output label space~\citep{pmlr-v162-chiu22a}.
Our vision and text models use an encoder-decoder architecture similar to T5 but incorporate a few changes.
Following PaLM~\citep{chowdhery2022palm}, we use multi-query attention~\citep{shazeer2019fast} that shares the same key and value projection for multiple query heads.
We only use 6 decoder layers and increase the feed forward hidden size (to compensate for the decrease in the number of layers).
These modifications have a neutral effect on model quality, but speed up auto-regressive decoding significantly.
We will show \name is compatible with these changes and can further speed up inference by a considerably large factor.
%For completeness, we also pretrain and evaluate text-only models using the modified model architecture.
We provide more details of our experimental setup in Appendix~\ref{appendix:exp_details}.

\begin{figure}[!t]
    \centering
    \includegraphics[width=5.0in]{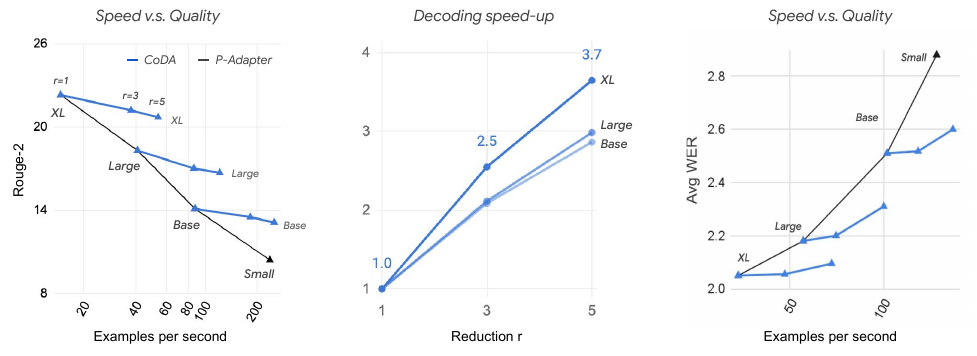}
    %\vspace{-0.1in}
    \caption{The scaling of \name on the XSum and LibriSpeech dataset. Left: \name achieves better speed-quality trade-off than finetuning adapters with smaller models, on the XSum dataset.
    Middle: larger \name model achieves higher speed-ups. Right: \name achieves better speed-quality trade-off than the dense baseline on the LibriSpeech dataset.} 
    \label{fig:nlp_scaling}
\end{figure}

\paragraph{NLP results}
In addition to the classification datasets used in Section~\ref{sec:ablation_results}, we also evaluate our final models on the SQuAD, ReCord and XSum datasets which require generating an answer or a summary given the input.
Table~\ref{table:nlp_finetuning_results} contains the finetuning results of XL models.
Compared to the parallel adapter baseline that uses full computation,
\name achieves 3x and 5x computation reduction with only 1.0 and 1.7 point loss in average score.
%The loss in model quality is higher on the three generation tasks, suggesting that a more difficult task requires more computation to process the token representations.

Figure~\ref{fig:avg_ft_nlp} and \ref{fig:nlp_scaling} highlight the scaling trend of \name.
\name runs much faster with slightly worse quality than the parallel adapter baseline.
This is expected because the baseline processes all tokens in every layer, whereas \name only selects $1/r$ of tokens for heavy processing.
Importantly, this quality gap reduces as the model size increases (as shown in Figure~\ref{fig:avg_ft_nlp}), making \name a computationally efficient choice for large models.
Indeed, \name can trade off quality for speed by varying the number of selected tokens.
Figure~\ref{fig:nlp_scaling} (left) demonstrates that \name achieves much stronger speed-quality trade-off compared to dense models without conditional computation.
The black line indicates the results of Parallel Adapter when the model size grows from Small to XL, and each blue line represents the speed-quality trade-off of \name using $r=1, 3, 5$.
Moreover, Figure~\ref{fig:nlp_scaling} (middle) shows that larger \name models exhibit higher inference speed-ups.
These observations are consistent on other tasks.
We provide additional results in Appendix~\S\ref{appendix:additional_results}.

%\begin{table}[ht]
%\centering
%\resizebox{3.255in}{!}{
%\begin{tabular}{lc@{~~~~}cccccccc}
%\toprule
%\multirow{2}{*}{Model} & \multirow{2}{*}{$r$} 
%& \multicolumn{2}{c}{Base} & & \multicolumn{2}{c}{Large} & & \multicolumn{2}{c}{XL} \\
%%& \multirow{2}{*}{$\Delta$Avg} \\
%\cmidrule{3-4} \cmidrule{6-7} \cmidrule{9-10} 
% &  & clean & other & & clean & other & & clean & other \\ % & \\
%\midrule
%w2v-BERT & - & 1.8 & 3.6 & & 1.5 & 2.9 & & 1.5 & 2.9 \\ % & - \\
%%\midrule
%BEST-RQ & - & 1.7 & 3.5 & & 1.6 & 2.9 & & 1.4 & 2.7 \\ % & - \\
%\midrule
%P-Adapter & - & 1.6 & 3.5 & & 1.4 & 3.0 & & 1.4 & 2.7 \\ % & $\pm$0.0 \\
%%\midrule[1pt]
%\name & \multirow{1}{*}{2} & 1.6 & 3.5 & & 1.4 & 3.0 & & 1.4 & 2.8 \\ % & \bf $+$0.02 \\
%%\midrule
%\name & \multirow{1}{*}{4} & 1.6 & 3.6 & & 1.5 & 3.1 & & 1.4 & 2.8 \\ % & \bf $+$0.07 \\
%\bottomrule
%\end{tabular}
%}
%\caption{Comparison of \name and the parallel adapter baselines on Librispeech. We report the WER results on test-clean and test-other. More results can be found in \S\ref{appendix:addition_speech_results}. Note that even with reduction $r=4$, the average performance drop 0.07 is minor.}
%\label{table:coda_speech}
%\end{table}

\begin{figure*}[!t]
\begin{floatrow}
\capbtabbox[.47\textwidth]{
    \resizebox{\linewidth}{!}{
        \centering
        \begin{tabular}{lc@{~~~~}cccccccc}
        \toprule
        \multirow{2}{*}{Model} & \multirow{2}{*}{$r$} 
        & \multicolumn{2}{c}{Base} & & \multicolumn{2}{c}{Large} & & \multicolumn{2}{c}{XL} \\
        %& \multirow{2}{*}{$\Delta$Avg} \\
        \cmidrule{3-4} \cmidrule{6-7} \cmidrule{9-10} 
         &  & clean & other & & clean & other & & clean & other \\ % & \\
        \midrule
        w2v-BERT & - & 1.8 & 3.6 & & 1.5 & 2.9 & & 1.5 & 2.9 \\ % & - \\
        %\midrule
        BEST-RQ & - & 1.7 & 3.5 & & 1.6 & 2.9 & & 1.4 & 2.7 \\ % & - \\
        \midrule
        P-Adapter & - & 1.6 & 3.5 & & 1.4 & 3.0 & & 1.4 & 2.7 \\ % & $\pm$0.0 \\
        %\midrule[1pt]
        \name & \multirow{1}{*}{2} & 1.6 & 3.5 & & 1.4 & 3.0 & & 1.4 & 2.8 \\ % & \bf $+$0.02 \\
        %\midrule
        \name & \multirow{1}{*}{4} & 1.6 & 3.6 & & 1.5 & 3.1 & & 1.4 & 2.8 \\ % & \bf $+$0.07 \\
        \bottomrule
        \end{tabular}
    }
    %\vspace{0.2in}
}{
    \caption{Comparison of \name and the parallel adapter baselines on Librispeech. We report the WER results on test-clean and test-other. More results can be found in \S\ref{appendix:addition_speech_results}.}
    %Note that even with reduction $r=4$, the average performance drop 0.07 is minor.}
    \label{table:coda_speech}
}
\capbtabbox[.49\textwidth]{
    \resizebox{.96\linewidth}{!}{
        \begin{tabular}{lcrrrrrr}
        \toprule
        \multirow{3}[2]{*}{\makecell{Model\\~}} &
        %\multirow{3}{*}{\makecell{Trainable\\Params}} &
        \multirow{3}{*}{$r$} &
        \multicolumn{2}{c}{OCRVQA} & 
        \multicolumn{2}{c}{DocVQA} &
        %\multicolumn{2}{c}{InfographicVQA} &
        \multicolumn{2}{c}{Screen2Words}\\
        \cmidrule(lr){3-4} \cmidrule(lr){5-6} \cmidrule(lr){7-8} 
        & & EM & Speedup & ANLS & Speedup  & CIDEr & Speedup\\
        \midrule
        \makecell{Parallel\\Adapter}  & - & 67.5 & 1$\times$ & 70.8 & 1$\times$ & 110.2 & 1$\times$ \\
        \cmidrule{1-8}
        \name & 4 & 68.2 & 4.6$\times$ & 71.8 & 4.6$\times$ & 111.6 & 4.6$\times$ \\
        \name & 8 & 67.6 & 8.0$\times$ & 66.6 & 8.0$\times$ & 108.1 & 8.0$\times$ \\
        \name & 16 & 66.9 & 13.5$\times$ & 56.6 & 12.1$\times$ & 109.0 & 12.5$\times$ \\
        \name & 32 & 64.4 & 19.4$\times$ & 42.5 & 16.7$\times$ & 104.2 & 17.8$\times$ \\
        \bottomrule
        \end{tabular}
    }
}{
\caption{Comparison of \name and the parallel adapter applied to a pretrained Pix2Struct model~\citep{lee2022pix2struct} on 3 visually-situated language understanding tasks.}
%We only compare between adapter methods to cleanly observe the effect of the conditional computation in \name.}
\label{tab:vision_results}
}
\end{floatrow}
\end{figure*}

\paragraph{Speech recognition results}
We further validate the performance of \name in the speech domain.
%, on a standard speech recognition dataset LibriSpeech.
Our model uses a Transformer encoder and a 2-layer LSTM Transducer \citep{graves2012sequence}.
%Our model adopts BERT pretraining with random quantized output labels, and is finetuned using supervised and noisy labels \citep{chiu2022self}. 
%The model architecture uses a Transformer encoder and a 2-layer LSTM Transducer \citep{graves2012sequence}. \name model is initialized from the pretrained models, with additional adapter and router parameters. Only adapters and routers are updated during finetuning. 
%CoDA learns to select $1/r$ of tokens from the heavy branch, which is complemented by the inexpensive adapter. Parallel adapter baselines simply set $r$=1 to use all the tokens for computation. 
Similar to NLP setups, we test the performance of the speech model on 3 scales -- Base, Large and XL (see Appendix~\ref{appendix:exp_details} for details). 
Table \ref{table:coda_speech} demonstrates that with sizable reduction ratios ($r=2,4$), the change on word error rate (WER) is consistently minimal on the test-clean and test-other sets of LibriSpeech
%(as well as on dev-clean and dev-other from \S\ref{appendix:addition_speech_results})
across different model sizes (and on other sets in \S\ref{appendix:addition_speech_results}). 
Moreover, our results are comparable to the top-performing models, such as w2v-BERT \citep{chung2021w2v} and BEST-RQ \citep{pmlr-v162-chiu22a}, that are fully finetuned by updating all parameters.
%\tl{Junwen: could you double check my revision and add the citations?}
%$\Delta$Avg measures the average difference in WER. From the table, both 0.03\% and 0.1\% are small. 
Figure~\ref{fig:nlp_scaling} (right) highlight again that applying conditional computation leads to better speed-quality trade-off compared to dense models.

%Figure~\ref{fig:speech_scaling} (left) shows the average WERs with regard to different scales.
%Again, the black line shows the speed-quality curve for the dense parallel adapter method and each blue line presents the trade-offs using $r=1,2,4$ at a given model size.
%\name can substantially increase the processing capability (i.e. examples per second). 
%More importantly, larger \name models are often faster and better than smaller parallel adapter baselines. For instance, XL model with $r=4$ has relatively 5\% lower average WER and runs 24\% faster than the Large parallel adapter baseline. Note that the XL model is almost two times larger than the Large model in size, but has an even faster speed with CoDA applied. Similarly, Base model with $r=4$ gives relatively 10\% better WER with 13\% faster speed than the Small parallel adapter model. Similar findings are also illustrated in Figure \ref{fig:speech_scaling} (right) showcasing the Giga floating-point operation count (GFLOPs) of different models. Larger \name models exhibit greater computation reduction.

\begin{figure*}[t!]
\resizebox{\columnwidth}{!}{
\centering
\begin{tabular*}{\textwidth}{cccc}
\textbf{Original Image} & \textbf{Layer 0} & \textbf{Layer 8} & \textbf{Layer 17}\\
\frame{\includegraphics[width=0.225\textwidth]{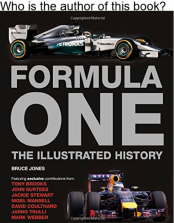}}
& 
\frame{\includegraphics[width=0.225\textwidth]{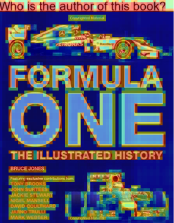}}
& 
\frame{\includegraphics[width=0.225\textwidth]{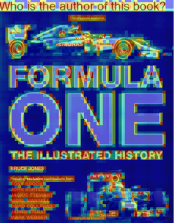}}
& 
\frame{\includegraphics[width=0.225\textwidth]{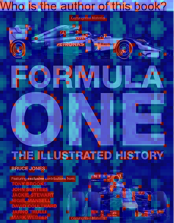}}
\end{tabular*}
\caption{Visualization of routing preferences for a \name model applied to the OCR-VQA task. Warmer and cooler colors represent higher and lower scores respectively. Router prefers diverse coverage in early layers, but converges to selecting sparse and representative patches in later layers.}
\label{fig:visualization}
}
\end{figure*}

%\begin{table}[t!]
%\setlength{\tabcolsep}{5pt}
%\centering
%\resizebox{3.2in}{!}{
%\begin{tabular}{lcrrrrrr}
%\toprule
%\multirow{3}[2]{*}{\makecell{Model\\~}} &
%%\multirow{3}{*}{\makecell{Trainable\\Params}} &
%\multirow{3}[2]{*}{\makecell{Reduction\\$r$\\~}} &
%\multicolumn{2}{c}{OCRVQA} & 
%\multicolumn{2}{c}{DocVQA} &
%%\multicolumn{2}{c}{InfographicVQA} &
%\multicolumn{2}{c}{Screen2Words}\\
%\cmidrule(lr){3-4} \cmidrule(lr){5-6} \cmidrule(lr){7-8} 
%& & EM & Speedup & ANLS & Speedup  & CIDEr & Speedup\\
%\midrule
%\makecell{Parallel\\Adapter}  & - & 67.5 & 1$\times$ & 70.8 & 1$\times$ & 110.2 & 1$\times$ \\
%\cmidrule{1-8}
%\name & 4 & 68.2 & 4.6$\times$ & 71.8 & 4.6$\times$ & 111.6 & 4.6$\times$ \\
%\name & 8 & 67.6 & 8.0$\times$ & 66.6 & 8.0$\times$ & 108.1 & 8.0$\times$ \\
%\name & 16 & 66.9 & 13.5$\times$ & 56.6 & 12.1$\times$ & 109.0 & 12.5$\times$ \\
%\name & 32 & 64.4 & 19.4$\times$ & 42.5 & 16.7$\times$ & 104.2 & 17.8$\times$ \\
%\bottomrule
%\end{tabular}
%}
%\caption{Comparison of \name and the parallel adapter applied to a pretrained Pix2Struct model~\citep{lee2022pix2struct} on 3 visually-situated language understanding tasks. We only compare between adapter methods to cleanly observe the effect of the conditional computation in \name.}
%\label{tab:vision_results}
%\end{table}

\paragraph{Vision results}

%We extend our experiments to the visual tasks that involve natural language within the image, such as documents, illustrations, and user interfaces. These tasks are particularly well suited for \name since long sequence length (for high resolution) is critical, and there is known sparsity of information in the underlying data. Our experiments are based on the Pix2Struct~\citep{lee2022pix2struct} approach, where an image-encoder-text-decoder is pretrained by learning to predict simplified HTML from web page screenshots.
%Aside from the minor architectural changes also applied in the NLP experiments to improve inference speed, we follow the pretraining procedure from the original paper exactly. Since our research question is on enabling a new type of parameter-efficient finetuning, we only perform head-to-head comparisons with the parallel adapter baseline.

We extend our experiments to visual tasks that involves natural language within the image, such as documents and user interfaces. Our experiments are based on Pix2Struct~\citep{lee2022pix2struct}, where an image-encoder-text-decoder is pretrained by learning to predict simplified HTML from webpage screenshots.
Table~\ref{tab:vision_results} shows the results on three tasks that were also evaluated in the original Pix2Struct paper. In OCRVQA and Screen2Words, we observe relatively small drops in performance when reducing the number of routed tokens (i.e. patches). When the capacity is 1/16th of the original sequence length, leading to around 13$\times$ speedup, we only lose about 1 point. We speculate that this is due to the high-level sparsity in the inputs for these two tasks. For DocVQA, where there is comparatively more textual information, we observe a steeper performance-speed trade-off but still achieve a 8$\times$ speedup with a 4-point drop.

To provide a more intuitive understanding why \name works, we visualize the router behavior for the OCR-VQA model in Figure~\ref{fig:visualization}. We show which patches the routers prefers the most (warmest colors) and least (coolest colors), for several layers. The first, immediately obvious, observation is that router avoids low-frequency patches, i.e. patches likely to be ``whitespace'', since they can be adequately handled by the cheap adapter layers. The second, more subtle, observation is that the router progressively converges on a small number of key patches that we hypothesize serve as representations for larger regions.
The visualization confirms that \name is able to select meaningful and representative patches that are useful for the prediction task.

%% file: sections/conclusion.tex
\section{Conclusion and Limitation}

We present \name, a parameter-efficient adapter method that enables fast inference.
\name relies on conditional computation to selectively activate model computation on important input units, providing a novel way to balance model expressivity and efficiency.

In this work, we focus on encoder-heavy applications such as summarization, speech recognition and visual question answering, by applying our method to the encoder.
One limitation of \name is that the current routing mechanism (i.e. token selection in a given sequence) is not directly applicable to decoder-only models for auto-regressive token generation.
Enabling fast token generation using conditional activation in decoder layers is an interesting direction we plan to explore in future work.

%Though the current routing mechanism (i.e. token selection in a given sequence) of \name is not directly applicable to decoder-only models for auto-regressive token generation, we will explore to extend our method to decoder models as future work, and research new variants of adaptive and conditional computation.

\section{Acknowledgements}
We would like to thank Rama Pasumarthi, Hongkun Yu, Kelvin Guu, Zhuyun Dai, Timothy Dozat, Raphael Hoffmann, Tao Wang, Tal Schuster, Ziwei Ji, Frederick Liu and Slav Petrov for helpful advice and discussion.

%% file: sections/appendix.tex
\section{Experimental details}
\label{appendix:exp_details}

\paragraph{Model implementation}
For our text and vision experiments, we implement our models using JAX~\citep{jax2018github}.
Specifically, our training and model modules are built on top of the T5X, Flax and Flaxformer framework~\citep{roberts2022t5x,flax2020github}.
Following the T5 v1.1 implementation and PaLM~\citep{chowdhery2022palm}, our Transformer models use the GLU variant~\citep{shazeer2020glu} as the feed forward network and multi-query-attention~\citep{shazeer2019fast} as the attention block.
These modifications are shown to enhance modeling capacity and speed up decoding respectively.

For the speech experiments, we use TensorFlow~\citep{tensorflow2015-whitepaper} and the Lingvo framework~\citep{shen2019lingvo}.
The state-of-the-art Transformer variant for speech recognition is the Conformer architecture~\citep{gulati2020conformer} which additionally uses depth-wise convolution in each layer. 
Since the convolution operation is applied to consecutive inputs and does not immediately support routing, we use the standard Transformer architecture~\citep{vaswani2017attention} instead. 
Swish activation is used in the feed forward blocks, following \citet{gulati2020conformer}.
We provide the model configuration details in Table~\ref{tab:model_config}.

\begin{table*}[th]
\footnotesize
    \centering
    \begin{tabular}{lccccccc}
    \toprule
    Model & Num of params & Layers & Num of heads & $d_\textrm{model}$ & $d_\textrm{ffn}$ & $d_\textrm{head}$ & $d_\textrm{adpt}$ \\
    \midrule
    Text Base & 0.1B & 12, 6 & 12 & 768 & 3072 & 128 & 64\\
    Text Large & 0.5B & 24, 6 & 16 & 1024 & 4096 & 128 & 64\\
    Text XL & 2.1B & 24, 6 & 32 & 2048 & 8192 & 128 & 64 \\
    \midrule
%    Speech Small & 0.1B & 30, 2 & 8 & 512 & 2048 & 64 & - \\
    Speech Base & 0.2B & 31, 2 & 8 & 768 & 3072 & 96  & 256\\
    Speech Large & 0.6B & 32, 2 & 8 & 1280 & 5120 & 160  & 256\\
    Speech XL & 1.1B & 32, 2 & 8 & 1664 & 6656 & 208 & 256\\
    \midrule 
    Vision & 0.7B & 18, 6 & 24 & 1536 & 3968 & 128 & 256 \\
    \bottomrule
    \end{tabular}
    \caption{Configuration of Transformer models used in \S\ref{sec:full_results}. We show the total number of parameters (in billions). $d_\textrm{model}$ is the model hidden dimension, $d_\textrm{ffn}$ is the intermediate FFN hidden dimension, $d_\textrm{head}$ is the attention head dimension and $d_\textrm{adpt}$ is the adapter hidden dimension.
    \label{tab:model_config}}
\end{table*}
\begin{table*}[th]
\footnotesize
\centering
\begin{tabular}{lccccc}
\toprule
Dataset & Input length & Batch size & Steps & Optimizer & Learning rate \\
\midrule
MNLI & 128 & \multirow{6}{*}{128} & \multirow{6}{*}{300K} & \multirow{6}{*}{Adafactor} & \multirow{6}{*}{0.001, constant} \\
RTE & 256 & & & & \\
BoolQ & 384 & \\
SQuAD & 512 & \\
ReCord & 512 & \\
XSum & 1664 & \\
\midrule
LibriSpeech & 3200 & 256 & 150K & Adafactor & 0.001, inverse decay \\
\midrule 
OCR-VQA & 4096 & 256 & \multirow{3}{*}{20K} & \multirow{3}{*}{Adafactor} & \multirow{3}{*}{0.01, cosine decay} \\
DocVQA & 4096 & 256 \\
%InforgraphicVQA & 6144 & 64 \\
Screen2Words & 4096 & 32 \\
\bottomrule
\end{tabular}
\caption{Fine-tuning hyperparmaeters. We use a maximum of 300K steps for NLP tasks following T5~\citep{t5}. Pix2struct~\citep{lee2022pix2struct} uses 10K fine-tuning steps for vision tasks. We use 20K steps as \name takes longer to train.}
\label{tab:finetuning_config}
\end{table*}

\paragraph{Model training}

We use the same data and procedure described in T5~\citep{t5}, BEST-RQ~\citep{pmlr-v162-chiu22a} and Pix2struct~\citep{lee2022pix2struct} for pre-training the respective text, speech and vision models.
We use the same training hyper-parameters, such as batch size, input sequence length, the number of pre-training steps and the choice of optimizer and learning rate scheduling. 
All models have been pre-trained using 128 or 256 TPUv3/TPUv4 chips.

We run CODA pre-training for text and vision models, using an additional 100K steps and 200K steps respectively.
For text models, the input sequence length is $n=512$ and we set the number of selected tokens $k=192$.
For vision models, the input sequence contains $n=4096$ image patches and we set $k=1024$.
CODA pre-training is not used for our speech models because there are sufficient fine-tuning data.
Following standard practice in speech, we use the 1K hour data from the LibriSpeech dataset \citep{panayotov2015librispeech} and another 30K hour data generated using the noisy student self-training method~\citep{xie2020self,zhang2022bigssl}.

Table~\ref{tab:finetuning_config} lists the hyper-parameters used for fine-tuning, including the sequence length, learning rate, batch size and the number of fine-tuning steps used. 
For NLP datasets, we set the maximum input length and decoding length to the 98th percentile of lengths in the training set.
For vision datasets, we set the input length following the suggested values in Pix2struct.
We also find that annealing the number of routed tokens $k$ can achieve better finetuning results. 
Specifically, we decrease $k$ linearly from the sequence length $n$ down to the target value $n/r$ using the first $10\%$ to $20\%$ of the finetuning steps.

\section{Additional results}
\label{appendix:additional_results}

\begin{figure*}[t!]
\centering
\begin{tabular}{c}
\includegraphics[width=5.2in]{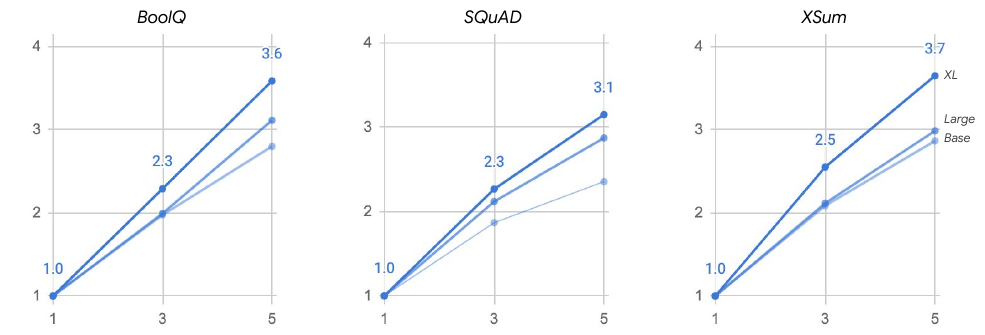} \\
(a) \\[0.5em]
\includegraphics[width=5.3in]{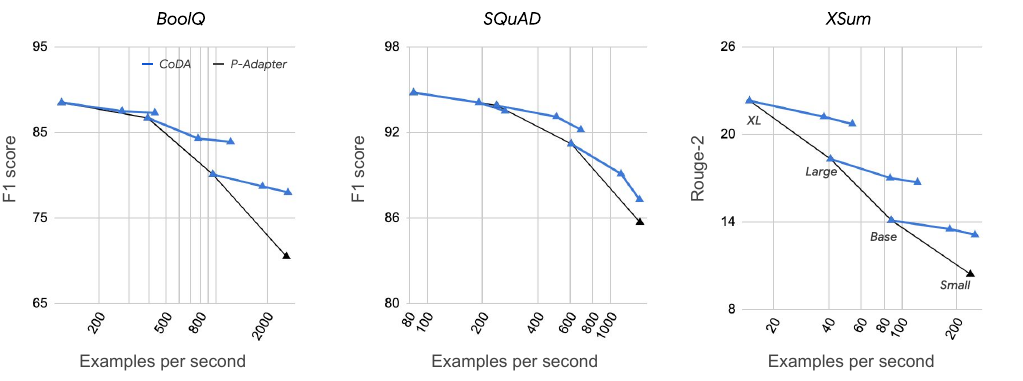} \\
(b) 
\end{tabular}
\caption{Analyzing the speed and quality of \name. 
We select 3 representative tasks including BoolQ (classification), SQuAD (question answering) and XSum (summarization).
(a) Relative decoding speed given different reduction factor $r$.
(b) Speed-quality trade-off of \name applied on different tasks and model sizes. \name achieves better quality than the dense Parallel Adapter baseline when running at a similar inference speed. The black line shows the performance of Parallel Adapter given Small to XL model size. Each blue line represents the performance of \name for reduction $r=1, 3, 5$ for a given model size. When $r=1$, \name is equivalent to the dense baseline.
\label{fig:nlp_speedup}}
\end{figure*}

\begin{table*}[!t]
\centering
\resizebox{5.4in}{!}{
\begin{tabular}{lccccccccc}
\toprule
\multirow{2}{*}{Model} & Trainable & Reduction & & $\,$MNLI$\,$ & $\,$RTE$\,$ & BoolQ & SQuAD & ReCord & $\,$XSum$\,$ \\
& Params & $r$ & & Acc. & Acc. & Acc. & F1 & F1 & R2 \\
\midrule
 & & & & \multicolumn{6}{c}{Base} \\
\cmidrule{1-3} \cmidrule{5-10} 
Parallel Adapter & \multirow{3}{*}{2M} & - & & 88.2 & 75.8 & 80.1 & 91.2 & 76.7 & 14.1 \\
\name & & 3 & & 85.8 & 68.6 & 78.7 & 89.1 & 69.1 & 13.5 \\
\name & & 5 & & 82.8 & 60.3 & 78.0 & 87.3 & 61.8 & 13.1 \\
\midrule
 & & & & \multicolumn{6}{c}{Large} \\
\cmidrule{1-3} \cmidrule{5-10} 
Parallel Adapter & \multirow{3}{*}{5M} & - & & 90.5 & 90.6 & 86.7 & 93.9 & 87.2 & 18.3 \\
\name & & 3 & & 90.0 & 84.8 & 84.3 & 93.1 & 84.6 & 17.0 \\
\name & & 5 & & 89.4 & 88.5 & 83.9 & 92.2 & 81.2 & 16.7  \\
\midrule
 & & & & \multicolumn{6}{c}{XL} \\
\cmidrule{1-3} \cmidrule{5-10} 
Parallel Adapter & \multirow{3}{*}{10M} & - & & 91.5 & 91.0 & 88.5 & 94.8 & 91.4 & 22.3 \\
\name & & 3 & & 91.2 & 90.3 & 87.5 & 94.1 & 89.3 & 21.2 \\
\name & & 5 & & 90.7 & 89.5 & 87.3 & 93.5 & 87.6 & 20.7  \\
\bottomrule
\end{tabular}
}
\caption{Fine-tuning results on 6 language tasks $\times$ 3 model sizes. We report the best results on the development set.}
\label{tab:full_nlp_finetuning_results}
\end{table*}

\subsection{NLP}
Table~\ref{tab:full_nlp_finetuning_results} contains the complete fine-tuning results on the 6 language datasets.
As discussed in \S\ref{sec:full_results}, the gap between \name and its counterpart without conditional computation is large at Base size.
As the model size increases, \name retains almost the same level of quality given 3x computation reduction ($r=3$).
The reduction leads to decoding speed-ups, as shown in Figure~\ref{fig:nlp_speedup}.
More importantly, we see that larger model benefits more from \name, achieving a speed-up factor close to the reduction factor $r$.
These results highlight the potential of \name for large-scale models, which we plan to investigate in future work.

\begin{table*}[t!]
\centering
\resizebox{5.4in}{!}{
\begin{tabular}{lc@{~~~~}cccccccccccccccccc}
\toprule
\multirow{3}{*}{Model} & \multirow{3}{*}{$r$} 
& \multicolumn{5}{c}{Base} & & \multicolumn{5}{c}{Large} & & \multicolumn{5}{c}{XL} & \multirow{3}{*}{$\Delta$ Avg} \\
%\multirow{2}{*}{Avg $\Delta$} \\
 \cmidrule{3-7} \cmidrule{9-13} \cmidrule{15-19}
 &  & \multicolumn{2}{c}{dev} & & \multicolumn{2}{c}{test} & & \multicolumn{2}{c}{dev} & & \multicolumn{2}{c}{test} & & \multicolumn{2}{c}{dev} & & \multicolumn{2}{c}{test} &  \\
\cmidrule{3-4} \cmidrule{6-7} \cmidrule{9-10} \cmidrule{12-13} \cmidrule{15-16} \cmidrule{18-19}
 &  & clean & other & & clean & other & & clean & other & & clean & other & & clean & other & & clean & other &  \\
\midrule
w2v-BERT  & - & 1.7 & 3.6 & & 1.8 & 3.6 & & 1.5 & 2.9 & & 1.5 & 2.9 & & 1.5 & 2.6 & & 1.5 & 2.9 &  - \\
BEST-RQ  & - & 1.6 & 3.5 & & 1.7 & 3.5 & & 1.5 & 2.8 & & 1.6 & 2.9 & & 1.4 & 2.7 & & 1.4 & 2.7 & - \\
\midrule
Parallel Adapter  & - & 1.5 & 3.5 & & 1.6 & 3.5 & & 1.4 & 3.0 & & 1.4 & 3.0 & & 1.4 & 2.7 &  & 1.4 & 2.7 & $\pm$0.0 \\
\name & \multirow{1}{*}{2} & 1.5 & 3.5 & & 1.6 & 3.5 & & 1.4 & 3.0 & & 1.4 & 3.0 & & 1.4 & 2.7 & & 1.4 & 2.8 & \bf $+$0.01 \\
\name & \multirow{1}{*}{4} & 1.6 & 3.6 & & 1.6 & 3.6 & & 1.4 & 3.1 & & 1.5 & 3.1 & & 1.4 & 2.8 & & 1.4 & 2.8 & \bf $+$0.07 \\
\bottomrule
\end{tabular}
}
\caption{Comparison of CoDA and the parallel adapter baselines on all 4 splits (dev-clean, dev-other, test-clean, test-other) of Librispeech.}
%The $\Delta$Avg numbers are even smaller than the ones from Table \ref{table:coda_speech}.}
\label{table:coda_speech_full}
\end{table*}

\subsection{Speech}
\label{appendix:addition_speech_results}

% SPEECH
\begin{figure}[t]
    \includegraphics[width=3.5in]{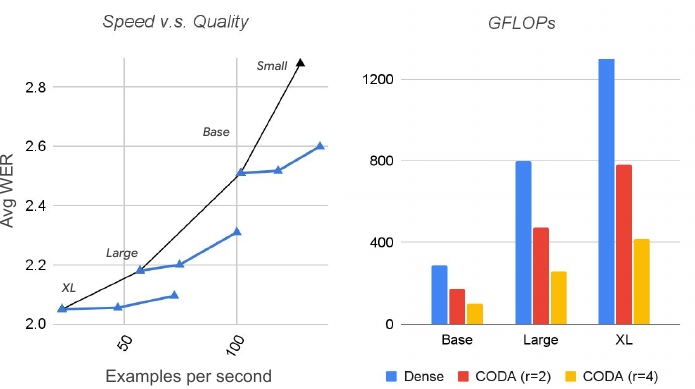}
  \caption{The scaling of CoDA on Librispeech. Left: Larger models with CoDA can achieve much better performance with the same or faster speed compared to smaller models. Right: GFLOPS reduction is significant as $r$ increases.}
  \label{fig:speech_scaling}
\end{figure}

Table \ref{table:coda_speech_full} extends Table \ref{table:coda_speech} by including WER results on dev-clean and dev-other splits. From the table, one can observe that XL with CoDA ($r=2,4$) are consistently better than the Large parallel adapter model on each split, and the Large model with CoDA ($r=2,4$) are also consistently better than the Base PA on each split. Given the inference speeds for CoDA models shown in Table \ref{fig:speech_scaling}, larger CoDA models are generally faster and better than smaller dense ones (even with PA) with regard to either time cost or computation GFLOPs. Therefore, it is likely for CoDA to help scale up ASR models with decent computation resources and time cost.

\subsection{Combining \name and LoRA}
\label{appendix:lora_results}
\name can be easily combined with other types of adapter methods.
To see this, we additionally implemented a variant that combines with Low-Rank Adapter~\citep[LoRA;][]{hu2021lora}, which is another parameter-efficient transfer learning method that recently became the most popular choice for LLMs. 
We incorporate the latest development suggested in the QLoRA paper~\citep{dettmers2023qlora}, which adds low-rank adapters to every linear projection matrix in the Transformer layers. 
This is found to obtain better fine-tuning performance than the original implementation.
Our \name variant with LoRA simply removes the parallel adapter branches and instead adds low-rank adapters to the projection matrices of the pretrained layers.

Table~\ref{table:coda_lora} shows the finetuning results.
The new LoRA baseline achieves stronger accuracy than the Parallel Adapter baseline (84.0 v.s. 82.9 on average), highlighting the effectiveness of recent development on LoRA.
In addition, our \name variant using LoRA still achieves very close accuracy compared to its dense counterpart (84.0 v.s. 84.0 or 83.7 on average).
We believe the additional results strengthen our claims -- that \name enables a strong trade-off between accuracy and efficiency using conditional activation, and this technique can be combined with other developments in PETL.

\begin{table*}[!t]

\centering
\resizebox{5.5in}{!}{
\begin{tabular}{lc@{~~~~}cccccccc}
\toprule
%\multirow{2}{*}{Model} & \multirow{2}{*}{Reduction $r$} 
& & \multicolumn{3}{c}{Base} & & \multicolumn{3}{c}{Large} & \\
%\multirow{2}{*}{Avg $\Delta$} \\
 \cmidrule{3-5} \cmidrule{7-9}
Model & Reduction $r\ $ & MNLI & RTE & BoolQ & & MNLI & RTE & BoolQ & Avg on 6 \\
\midrule
Parallel Adapter (PA)  & - & 87.1 & 71.5 & 77.9 & & 90.3 & 84.8 & 85.8 & 82.9\\
\name w/ PA & 3 & 86.6 & 72.6 & 76.6 & & 90.2 & 85.9 & 85.1 & 82.8 \\
\name w/ PA & 5 & 86.0 & 70.8 & 76.0 & & 89.7 & 85.2 & 84.3 & 82.0 \\
\midrule
Low-rank Adapter (LoRA) & - & 88.0 & 73.7 & 80.3 & & 90.7 & 85.2 & 86.3 & 84.0 \\
\name w/ LoRA & 3 & 86.2 & 76.9 & 78.4 & & 90.3 & 86.3 & 85.8 & 84.0 \\
\name w/ LoRA & 5 & 86.0 & 76.9 & 78.3 & & 89.8 & 86.3 & 84.7 & 83.7 \\
\bottomrule
\end{tabular}
}
\caption{Results of applying \name with Parallel Adapter v.s. Low-Rank Adapter. We report results on the development sets and on the Base and Large model size.}
\label{table:coda_lora}
\end{table*}

\section{Soft top-$k$ algorithm}
\label{appendix:soft_topk}

\subsection{Derivation of the iterative updates}
We present the derivation of iterative updates (\ref{eq:iterations}) for solving the soft top-$k$ problem~(\ref{eq:soft_topk}) in Section~\ref{sec:method}.
The soft top-$k$ operation is defined as a maximization problem~(\ref{eq:soft_topk}).
For the derivation, we rewrite it as an equivalent minimization problem:
\begin{align}
 & \max_{\bm{\lambda}} \ \ \vs^\top \bm{\lambda} + \epsilon H(\bm{\lambda}) \nonumber \\
\Longleftrightarrow\qquad & \min_{\bm{\lambda}} \ \ -\vs^\top \bm{\lambda} - \epsilon H(\bm{\lambda}) \nonumber \\
\Longleftrightarrow\qquad & \min_{\bm{\lambda}} \ \ -\vs^\top \bm{\lambda} - \epsilon H(\bm{\lambda}) - \epsilon \bm{1}^\top \bm{\lambda} \label{eq:minobj}\\
\text{s.t.}\  & \bm{1}^\top\bm{\lambda} = k, \ \  \bm{\lambda[i]}\in [0, 1],\ \ i=1,\dots,n. \nonumber
\end{align}
Note the term $\epsilon \bm{1}^\top \bm{\lambda}$ will be a constant $\epsilon\times k$, but we include it in the minimization object to make our derivation simpler later.

Now, let $a\in\R$ and $\vb\in\R^m$ be the Lagrangian variables corresponding to the linear constraints $\bm{1}^\top\bm{\lambda} = k$ and $\bm{\lambda}[i] \leq 1\ \forall i\ $.\footnote{$\bm{\lambda}[i] \geq 0\ \forall i\ $ is already implied by the term $H(\bm{\lambda}) = \sum_i -\bm{\lambda}[i] \log \bm{\lambda}[i]$ in the objective, due to the use of $\log\bm{\lambda}[i]$.}
The minimization problem is equivalent to its Lagrangian expression:
\begin{align}
\min_{\lambda\in\R^m}\ \max_{a\in\R, \vb\leq \bm{0}} \ -\vs^\top \bm{\lambda} - \epsilon H(\bm{\lambda}) - \epsilon \bm{1}^\top \bm{\lambda} + a (k-\bm{1}^\top\bm{\lambda}) + \bm{b}^\top (\bm{1} - \bm{\lambda})
\label{eq:minmax}
\end{align}
The objective function~(\ref{eq:minobj}) is strongly convex and the solution space of $\bm{\lambda}$ is a convex set. As a result, strong duality holds so we can instead solve the dual problem. The dual problem exchanges the $\min$ and $\max$ operators in (\ref{eq:minmax}):
\begin{align}
\max_{a\in\R, \vb\leq \bm{0}}\ \min_{\lambda\in\R^m}\ -\vs^\top \bm{\lambda} - \epsilon H(\bm{\lambda}) - \epsilon \bm{1}^\top \bm{\lambda} + a (k-\bm{1}^\top\bm{\lambda}) + \bm{b}^\top (\bm{1} - \bm{\lambda})
\label{eq:maxmin}
\end{align}
The optimal solution $(a, \vb, \bm{\lambda})$ must have the Karush-Kuhn-Tucker (KKT) conditions hold~\citep{kuhn2014nonlinear}, namely
\begin{align*}
& \frac{\partial\left(-\vs^\top \bm{\lambda} - \epsilon H(\bm{\lambda}) + \epsilon \bm{1}^\top \bm{\lambda} + a (k-\bm{1}^\top\bm{\lambda}) + \bm{b}^\top (\bm{1} - \bm{\lambda})\right)}{\partial \bm{\lambda}} = 0 \\
\Longleftrightarrow\qquad & \bm{\lambda} = \exp\left(\frac{\vs + a +\vb}{\epsilon}\right)
\quad \Longleftrightarrow\quad \bm{\lambda}[i] = \exp\left(\frac{\vs[i] + a + \vb[i]}{\epsilon}\right) \ \ \forall i=1,\dots,n
%\Longleftrightarrow\qquad & \lambda_i = \exp\left(\frac{s_i + a+\vb_i}{\epsilon}\right) \ \ \forall i=1,\dots,m
\end{align*}
Substituting $\bm{\lambda}$ using the above equation in (\ref{eq:maxmin}), the dual problem now has a simple form:
\begin{align*}
\max_{a\in\R, \vb\leq \bm{0}} k\cdot a + \bm{1}^\top\vb - \bm{1}^\top \exp\left( \frac{\vs + a +\vb}{\epsilon}\right)
\end{align*}
We can solve this problem using coordinate descent~\citep{wright2015coordinate} by successively maximizing the function with either $a$ or $\vb$ fixed.
That is, we find the optimal $a$ that maximizes the dual objective given a fixed $\vb$, and vice versa.
This leads to the iterative updates (\ref{eq:iterations}) described in Section~\ref{sec:method}.
\begin{equation}
\begin{split}
    a' = \epsilon \ln(k) - \epsilon \ln\left(\sum_{i=1}^n \exp\left(\frac{\vs[i] + \vb[i]}{\epsilon}\right)\right), \qquad \qquad \vb' = \min(-\vs - a', 0) \nonumber
\end{split}
\end{equation}

In short, we obtain the iterative updates of the soft top-$k$ problem~(\ref{eq:soft_topk}) by solving its dual problem and by performing coordinate decent of the dual variables $a$ and $\vb$.
The iterative updates are in fact the coordinate decent steps.

\subsection{The $\epsilon$-scaling trick}
The iterations of $a$ and $\vb$ will converge but the number of iterations needed can be very large for small $\epsilon$.
In practice, we only perform a small number of iterations and return the corresponding $\bm{\lambda}$, which may be close but not the exact solution to (\ref{eq:minobj}).
In order to improve the convergence given a small number of iterations, we apply an empirical trick called the \emph{$\epsilon$-scaling heuristic}~\citep{schmitzer2019stabilized}.
Let $\epsilon_t$ denote the value of $\epsilon$ at the $t$-th iteration.
We initialize $\epsilon_0$ to a larger value and gradually reduce $\epsilon_t$ to the target $\epsilon$.
Specifically, we set $\epsilon_t = \max(\beta \epsilon_{t-1}, \epsilon)$ at the $t$-th iteration, using a scaling constant $\beta \in (0, 1)$.
We use $\epsilon_0=4$ throughout our experiments, $\epsilon=0.03$ and $\beta=0.7$ for text and vision models and $\epsilon=1.0$ and $\beta=0.85$ for speech models.
Using a larger number of iterations leads to better convergence but we found $T=20$ sufficient for our experiments.

\subsection{Overhead of soft top-$k$ iterations}
The soft top-$k$ iterations are performed for every routed Transformer layer.
Although this seems computationally expensive, the actual overhead is very small compared to the overall decoding latency. 
The complexity only scales linearly with the number of layers and the sequence length, and does not depend on the model dimension $d$.
Table~\ref{table:iteration_latency} showcases the latency numbers on the BoolQ and XSum datasets, when performing batched decoding using a single TPUv4 chip.
We observe that the cost of iterations is less than 2\% of the total decoding latency.
Moreover, the relative cost decreases dramatically as the model size increases, since it does not depend on the model dimension.

\begin{table*}[th!]
\small
\centering
\begin{tabular}{lccc}
\toprule
Task &  Model size & Iteration latency (ms) & Total latency (ms)\\
\midrule
\multirow{3}{*}{BoolQ} & Base & 0.9 & 48.0 \\
& Large & 1.7 & 105.3 \\
& XL & 1.7 & 297.6 \\
\midrule
\multirow{3}{*}{XSum} & Base & 1.0 & 513.9 \\
& Large & 2.0 & 1076.2 \\
& XL & 2.0 & 2426.0 \\
\bottomrule
\end{tabular}
\caption{Latency (in milliseconds) of the soft top-$k$ iteration and the total decoding time per batch. We use a single TPUv4 chip and 128 sequences per batch. The maximum iteration overhead is less than 2\% of the total latency.}
\label{table:iteration_latency}
\end{table*}

\subsection{Additional discussion}
This iterative algorithm is closely related to the Sinkhorn algorithm of Optimal Transport (OT). 
Specifically, the Sinkhorn algorithm solves the entropy-regularized version of Optimal Transport~\citep{cuturi2013sinkhorn}.
However, our work concerns an different optimization instance. While OT solves a transportation problem where the solution space is defined with the marginal constraints over the rows and columns of a transportation matrix, our optimization problem is constrained with a total budget ($\sum_i \lambda_i = k$) and upper bounds ($\lambda_i \leq 1\ \forall i$). 
This leads to different iterative updates.

Concurrent to our work, \citet{tai2022spartan} have used a similar linear program (LP) formulation for soft top-$k$ operation, and have applied the operator for learning sparse neural networks (i.e. model pruning).
Compared to our formulation~(\ref{eq:minobj}), they first reduce the LP to an equivalent instance of optimal transport problem, before introducing the entropy term. 
As a result, the derived updates are different.
In addition, \citet{tai2022spartan} have introduced an initialization for the dual variables to improve the convergence of their algorithm, whereas we use $\epsilon$ scaling instead.
Their implementation can be explored for \name as well.

Besides formulating soft top-$k$ using entropy-regularized optimization, there are other possible variants for trainable sparsity.
One example is sparsemax~\citep{pmlr-v48-martins16} that can learn sparse multi-label probabilities.
We believe that the sparsemax formulation can generalize from the top-1 to top-$k$ case, but it is beyond the scope of this work. 
We use the current soft top-$k$ implementation because it is a natural extension of softmax (see discussions in \S\ref{sec:method}), and because it can be solved using simple iterative updates.

\section{Author Contributions}
All authors have contributed to running experiments and discussing research ideas.
Tao leads the project, developed the conditional architecture, designed the experiments and analyses.
Kenton, Yu and Ming-Wei proposed the idea of applying conditional computation for large model adaptation.
Joshua demonstrated the conditional architecture is applicable to attention, and implemented the initial version of conditional attention block.
Tao, Yanqi, Nan, Vincent, Yuexin, Ming-Wei and Yu conducted the NLP experiments including model pre-training, fine-tuning and various ablation analyses.
Siddhartha conducted the majority of the vision experiments.
Kenton conducted the vision analysis and advised on the vision experiments.
Junwen conducted the majority of the speech experiments.
Bo and Yu assisted in trouble-shooting the speech models, ran the model pre-training and provided guidance on the speech experiments.
Finally, Tao, Ming-Wei, Junwen and Kenton made the primary contributions to the writing of the paper.

%% file: main.bbl
\begin{thebibliography}{75}
\providecommand{\natexlab}[1]{#1}
\providecommand{\url}[1]{\texttt{#1}}
\expandafter\ifx\csname urlstyle\endcsname\relax
  \providecommand{\doi}[1]{doi: #1}\else
  \providecommand{\doi}{doi: \begingroup \urlstyle{rm}\Url}\fi

\bibitem[Abadi et~al.(2015)Abadi, Agarwal, Barham, Brevdo, Chen, Citro,
  Corrado, Davis, Dean, Devin, Ghemawat, Goodfellow, Harp, Irving, Isard, Jia,
  Jozefowicz, Kaiser, Kudlur, Levenberg, Man\'{e}, Monga, Moore, Murray, Olah,
  Schuster, Shlens, Steiner, Sutskever, Talwar, Tucker, Vanhoucke, Vasudevan,
  Vi\'{e}gas, Vinyals, Warden, Wattenberg, Wicke, Yu, and
  Zheng]{tensorflow2015-whitepaper}
Mart\'{i}n Abadi, Ashish Agarwal, Paul Barham, Eugene Brevdo, Zhifeng Chen,
  Craig Citro, Greg~S. Corrado, Andy Davis, Jeffrey Dean, Matthieu Devin,
  Sanjay Ghemawat, Ian Goodfellow, Andrew Harp, Geoffrey Irving, Michael Isard,
  Yangqing Jia, Rafal Jozefowicz, Lukasz Kaiser, Manjunath Kudlur, Josh
  Levenberg, Dandelion Man\'{e}, Rajat Monga, Sherry Moore, Derek Murray, Chris
  Olah, Mike Schuster, Jonathon Shlens, Benoit Steiner, Ilya Sutskever, Kunal
  Talwar, Paul Tucker, Vincent Vanhoucke, Vijay Vasudevan, Fernanda Vi\'{e}gas,
  Oriol Vinyals, Pete Warden, Martin Wattenberg, Martin Wicke, Yuan Yu, and
  Xiaoqiang Zheng.
\newblock {TensorFlow}: Large-scale machine learning on heterogeneous systems,
  2015.
\newblock URL \url{https://www.tensorflow.org/}.
\newblock Software available from tensorflow.org.

\bibitem[Ainslie et~al.(2023)Ainslie, Lei, de~Jong, Onta{\~n}{\'o}n, Brahma,
  Zemlyanskiy, Uthus, Guo, Lee-Thorp, Tay, et~al.]{ainslie2023colt5}
Joshua Ainslie, Tao Lei, Michiel de~Jong, Santiago Onta{\~n}{\'o}n, Siddhartha
  Brahma, Yury Zemlyanskiy, David Uthus, Mandy Guo, James Lee-Thorp, Yi~Tay,
  et~al.
\newblock Colt5: Faster long-range transformers with conditional computation.
\newblock In \emph{Proceedings of the 2023 Conference on Empirical Methods in
  Natural Language Processing (EMNLP)}, 2023.

\bibitem[Awadalla et~al.(2022)Awadalla, Wortsman, Ilharco, Min, Magnusson,
  Hajishirzi, and Schmidt]{awadalla2022exploring}
Anas Awadalla, Mitchell Wortsman, Gabriel Ilharco, Sewon Min, Ian Magnusson,
  Hannaneh Hajishirzi, and Ludwig Schmidt.
\newblock Exploring the landscape of distributional robustness for question
  answering models.
\newblock \emph{arXiv preprint arXiv:2210.12517}, 2022.

\bibitem[Ba et~al.(2016)Ba, Kiros, and Hinton]{ba2016layer}
Jimmy~Lei Ba, Jamie~Ryan Kiros, and Geoffrey~E Hinton.
\newblock Layer normalization.
\newblock \emph{arXiv preprint arXiv:1607.06450}, 2016.

\bibitem[Bapna et~al.(2020)Bapna, Arivazhagan, and Firat]{bapna2020controlling}
Ankur Bapna, Naveen Arivazhagan, and Orhan Firat.
\newblock Controlling computation versus quality for neural sequence models.
\newblock \emph{arXiv preprint arXiv:2002.07106}, 2020.

\bibitem[Beltagy et~al.(2020)Beltagy, Peters, and Cohan]{Beltagy2020Longformer}
Iz~Beltagy, Matthew~E. Peters, and Arman Cohan.
\newblock Longformer: The long-document transformer.
\newblock \emph{arXiv:2004.05150}, 2020.

\bibitem[Bentivogli et~al.(2009)Bentivogli, Clark, Dagan, and
  Giampiccolo]{bentivogli2009fifth}
Luisa Bentivogli, Peter Clark, Ido Dagan, and Danilo Giampiccolo.
\newblock The fifth pascal recognizing textual entailment challenge.
\newblock In \emph{TAC}, 2009.
\newblock URL
  \url{https://tac.nist.gov/publications/2009/additional.papers/RTE5_overview.proceedings.pdf}.

\bibitem[Bolya et~al.(2023)Bolya, Fu, Dai, Zhang, Feichtenhofer, and
  Hoffman]{bolya2023tome}
Daniel Bolya, Cheng-Yang Fu, Xiaoliang Dai, Peizhao Zhang, Christoph
  Feichtenhofer, and Judy Hoffman.
\newblock Token merging: Your {ViT} but faster.
\newblock In \emph{International Conference on Learning Representations}, 2023.

\bibitem[Bradbury et~al.(2018)Bradbury, Frostig, Hawkins, Johnson, Leary,
  Maclaurin, Necula, Paszke, VanderPlas, Wanderman-Milne,
  et~al.]{jax2018github}
James Bradbury, Roy Frostig, Peter Hawkins, Matthew~James Johnson, Chris Leary,
  Dougal Maclaurin, George Necula, Adam Paszke, Jake VanderPlas, Skye
  Wanderman-Milne, et~al.
\newblock Jax: composable transformations of python+ numpy programs.
\newblock \emph{Version 0.2}, 5:\penalty0 14--24, 2018.

\bibitem[Chiu et~al.(2022)Chiu, Qin, Zhang, Yu, and Wu]{pmlr-v162-chiu22a}
Chung-Cheng Chiu, James Qin, Yu~Zhang, Jiahui Yu, and Yonghui Wu.
\newblock Self-supervised learning with random-projection quantizer for speech
  recognition.
\newblock In Kamalika Chaudhuri, Stefanie Jegelka, Le~Song, Csaba Szepesvari,
  Gang Niu, and Sivan Sabato, editors, \emph{Proceedings of the 39th
  International Conference on Machine Learning}, volume 162 of
  \emph{Proceedings of Machine Learning Research}, pages 3915--3924. PMLR,
  17--23 Jul 2022.
\newblock URL \url{https://proceedings.mlr.press/v162/chiu22a.html}.

\bibitem[Chowdhery et~al.(2022)Chowdhery, Narang, Devlin, Bosma, Mishra,
  Roberts, Barham, Chung, Sutton, Gehrmann, et~al.]{chowdhery2022palm}
Aakanksha Chowdhery, Sharan Narang, Jacob Devlin, Maarten Bosma, Gaurav Mishra,
  Adam Roberts, Paul Barham, Hyung~Won Chung, Charles Sutton, Sebastian
  Gehrmann, et~al.
\newblock Palm: Scaling language modeling with pathways.
\newblock \emph{arXiv preprint arXiv:2204.02311}, 2022.

\bibitem[Chung et~al.(2021)Chung, Zhang, Han, Chiu, Qin, Pang, and
  Wu]{chung2021w2v}
Yu-An Chung, Yu~Zhang, Wei Han, Chung-Cheng Chiu, James Qin, Ruoming Pang, and
  Yonghui Wu.
\newblock W2v-bert: Combining contrastive learning and masked language modeling
  for self-supervised speech pre-training.
\newblock In \emph{2021 IEEE Automatic Speech Recognition and Understanding
  Workshop (ASRU)}, pages 244--250. IEEE, 2021.

\bibitem[Clark et~al.(2019)Clark, Lee, Chang, Kwiatkowski, Collins, and
  Toutanova]{clark-etal-2019-boolq}
Christopher Clark, Kenton Lee, Ming-Wei Chang, Tom Kwiatkowski, Michael
  Collins, and Kristina Toutanova.
\newblock {B}ool{Q}: Exploring the surprising difficulty of natural yes/no
  questions.
\newblock In \emph{Proceedings of the 2019 Conference of the North {A}merican
  Chapter of the Association for Computational Linguistics: Human Language
  Technologies, Volume 1 (Long and Short Papers)}, 2019.
\newblock URL \url{https://aclanthology.org/N19-1300}.

\bibitem[Cuturi(2013)]{cuturi2013sinkhorn}
Marco Cuturi.
\newblock Sinkhorn distances: Lightspeed computation of optimal transport.
\newblock \emph{Advances in neural information processing systems}, 26, 2013.

\bibitem[Dagan et~al.(2005)Dagan, Glickman, and Magnini]{Dagan:2005}
Ido Dagan, Oren Glickman, and Bernardo Magnini.
\newblock The pascal recognising textual entailment challenge.
\newblock In \emph{Proceedings of the PASCAL Challenges Workshop on Recognising
  Textual Entailment}, 2005.
\newblock URL \url{http://www.cs.biu.ac.il/~glikmao/rte05/}.

\bibitem[Dettmers et~al.(2023)Dettmers, Pagnoni, Holtzman, and
  Zettlemoyer]{dettmers2023qlora}
Tim Dettmers, Artidoro Pagnoni, Ari Holtzman, and Luke Zettlemoyer.
\newblock Qlora: Efficient finetuning of quantized llms.
\newblock \emph{arXiv preprint arXiv:2305.14314}, 2023.

\bibitem[Devlin et~al.(2019)Devlin, Chang, Lee, and Toutanova]{devlin2019bert}
Jacob Devlin, Ming-Wei Chang, Kenton Lee, and Kristina Toutanova.
\newblock Bert: Pre-training of deep bidirectional transformers for language
  understanding.
\newblock \emph{ArXiv}, abs/1810.04805, 2019.

\bibitem[Du et~al.(2022)Du, Huang, Dai, Tong, Lepikhin, Xu, Krikun, Zhou, Yu,
  Firat, et~al.]{du2022glam}
Nan Du, Yanping Huang, Andrew~M Dai, Simon Tong, Dmitry Lepikhin, Yuanzhong Xu,
  Maxim Krikun, Yanqi Zhou, Adams~Wei Yu, Orhan Firat, et~al.
\newblock Glam: Efficient scaling of language models with mixture-of-experts.
\newblock In \emph{International Conference on Machine Learning}, pages
  5547--5569. PMLR, 2022.

\bibitem[Fedus et~al.(2021)Fedus, Zoph, and Shazeer]{fedus2021switch}
William Fedus, Barret Zoph, and Noam Shazeer.
\newblock Switch transformers: Scaling to trillion parameter models with simple
  and efficient sparsity, 2021.

\bibitem[Giampiccolo et~al.(2007)Giampiccolo, Magnini, Dagan, and
  Dolan]{giampiccolo2007third}
Danilo Giampiccolo, Bernardo Magnini, Ido Dagan, and William~B Dolan.
\newblock The third pascal recognizing textual entailment challenge.
\newblock In \emph{Proceedings of the ACL-PASCAL workshop on textual entailment
  and paraphrasing}, 2007.
\newblock URL \url{https://aclanthology.org/W07-1401.pdf}.

\bibitem[Graves(2012)]{graves2012sequence}
Alex Graves.
\newblock Sequence transduction with recurrent neural networks.
\newblock \emph{arXiv preprint arXiv:1211.3711}, 2012.

\bibitem[Gulati et~al.(2020)Gulati, Qin, Chiu, Parmar, Zhang, Yu, Han, Wang,
  Zhang, Wu, et~al.]{gulati2020conformer}
Anmol Gulati, James Qin, Chung-Cheng Chiu, Niki Parmar, Yu~Zhang, Jiahui Yu,
  Wei Han, Shibo Wang, Zhengdong Zhang, Yonghui Wu, et~al.
\newblock Conformer: Convolution-augmented transformer for speech recognition.
\newblock \emph{arXiv preprint arXiv:2005.08100}, 2020.

\bibitem[Guo et~al.(2022)Guo, Ainslie, Uthus, Ontanon, Ni, Sung, and
  Yang]{guo-etal-2022-longt5}
Mandy Guo, Joshua Ainslie, David Uthus, Santiago Ontanon, Jianmo Ni, Yun-Hsuan
  Sung, and Yinfei Yang.
\newblock {L}ong{T}5: {E}fficient text-to-text transformer for long sequences.
\newblock In \emph{Findings of the Association for Computational Linguistics:
  NAACL 2022}. Association for Computational Linguistics, 2022.
\newblock URL \url{https://aclanthology.org/2022.findings-naacl.55}.

\bibitem[Haim et~al.(2006)Haim, Dagan, Dolan, Ferro, Giampiccolo, Magnini, and
  Szpektor]{haim2006second}
R~Bar Haim, Ido Dagan, Bill Dolan, Lisa Ferro, Danilo Giampiccolo, Bernardo
  Magnini, and Idan Szpektor.
\newblock The second pascal recognising textual entailment challenge.
\newblock In \emph{Proceedings of the Second PASCAL Challenges Workshop on
  Recognising Textual Entailment}, 2006.
\newblock URL \url{http://u.cs.biu.ac.il/~nlp/RTE2/Proceedings/01.pdf}.

\bibitem[Han et~al.(2016)Han, Mao, and Dally]{han2015deep_compression}
Song Han, Huizi Mao, and William~J Dally.
\newblock Deep compression: Compressing deep neural networks with pruning,
  trained quantization and huffman coding.
\newblock \emph{International Conference on Learning Representations (ICLR)},
  2016.

\bibitem[He et~al.(2021)He, Zhou, Ma, Berg-Kirkpatrick, and
  Neubig]{he2021towards}
Junxian He, Chunting Zhou, Xuezhe Ma, Taylor Berg-Kirkpatrick, and Graham
  Neubig.
\newblock Towards a unified view of parameter-efficient transfer learning.
\newblock \emph{arXiv preprint arXiv:2110.04366}, 2021.

\bibitem[Heek et~al.(2020)Heek, Levskaya, Oliver, Ritter, Rondepierre, Steiner,
  and van Zee]{flax2020github}
Jonathan Heek, Anselm Levskaya, Avital Oliver, Marvin Ritter, Bertrand
  Rondepierre, Andreas Steiner, and Marc van Zee.
\newblock Flax: A neural network library and ecosystem for jax.
\newblock \emph{Version 0.3}, 3:\penalty0 14--26, 2020.

\bibitem[Hinton et~al.(2015)Hinton, Vinyals, and Dean]{hinton2015distilling}
Geoffrey Hinton, Oriol Vinyals, and Jeff Dean.
\newblock Distilling the knowledge in a neural network.
\newblock \emph{arXiv preprint arXiv:1503.02531}, 2015.

\bibitem[Houlsby et~al.(2019)Houlsby, Giurgiu, Jastrzebski, Morrone,
  De~Laroussilhe, Gesmundo, Attariyan, and Gelly]{houlsby2019parameter}
Neil Houlsby, Andrei Giurgiu, Stanislaw Jastrzebski, Bruna Morrone, Quentin
  De~Laroussilhe, Andrea Gesmundo, Mona Attariyan, and Sylvain Gelly.
\newblock Parameter-efficient transfer learning for nlp.
\newblock In \emph{International Conference on Machine Learning}, pages
  2790--2799. PMLR, 2019.

\bibitem[Hu et~al.(2021)Hu, Shen, Wallis, Allen-Zhu, Li, Wang, and
  Chen]{hu2021lora}
Edward Hu, Yelong Shen, Phil Wallis, Zeyuan Allen-Zhu, Yuanzhi Li, Lu~Wang, and
  Weizhu Chen.
\newblock Lora: Low-rank adaptation of large language models, 2021.

\bibitem[Hua et~al.(2022)Hua, Dai, Liu, and Le]{hua2022transformer}
Weizhe Hua, Zihang Dai, Hanxiao Liu, and Quoc Le.
\newblock Transformer quality in linear time.
\newblock In \emph{International Conference on Machine Learning}, pages
  9099--9117. PMLR, 2022.

\bibitem[{Kahn} et~al.(2020){Kahn}, {Rivière}, {Zheng}, {Kharitonov}, {Xu},
  {Mazaré}, {Karadayi}, {Liptchinsky}, {Collobert}, {Fuegen}, {Likhomanenko},
  {Synnaeve}, {Joulin}, {Mohamed}, and {Dupoux}]{librilight}
J.~{Kahn}, M.~{Rivière}, W.~{Zheng}, E.~{Kharitonov}, Q.~{Xu}, P.~E.
  {Mazaré}, J.~{Karadayi}, V.~{Liptchinsky}, R.~{Collobert}, C.~{Fuegen},
  T.~{Likhomanenko}, G.~{Synnaeve}, A.~{Joulin}, A.~{Mohamed}, and E.~{Dupoux}.
\newblock Libri-light: A benchmark for asr with limited or no supervision.
\newblock In \emph{ICASSP 2020 - 2020 IEEE International Conference on
  Acoustics, Speech and Signal Processing (ICASSP)}, pages 7669--7673, 2020.

\bibitem[Kim and Rush(2016)]{kim-rush-2016-sequence}
Yoon Kim and Alexander~M. Rush.
\newblock Sequence-level knowledge distillation.
\newblock In \emph{Proceedings of the 2016 Conference on Empirical Methods in
  Natural Language Processing (EMNLP)}, 2016.
\newblock URL \url{https://aclanthology.org/D16-1139}.

\bibitem[Kuhn and Tucker(2014)]{kuhn2014nonlinear}
Harold~W Kuhn and Albert~W Tucker.
\newblock Nonlinear programming.
\newblock In \emph{Traces and emergence of nonlinear programming}, pages
  247--258. Springer, 2014.

\bibitem[Lee et~al.(2022)Lee, Joshi, Turc, Hu, Liu, Eisenschlos, Khandelwal,
  Shaw, Chang, and Toutanova]{lee2022pix2struct}
Kenton Lee, Mandar Joshi, Iulia Turc, Hexiang Hu, Fangyu Liu, Julian
  Eisenschlos, Urvashi Khandelwal, Peter Shaw, Ming-Wei Chang, and Kristina
  Toutanova.
\newblock Pix2struct: Screenshot parsing as pretraining for visual language
  understanding.
\newblock \emph{arXiv preprint arXiv:2210.03347}, 2022.

\bibitem[Lei(2021)]{lei-2021-attention}
Tao Lei.
\newblock When attention meets fast recurrence: Training language models with
  reduced compute.
\newblock In \emph{Proceedings of the 2021 Conference on Empirical Methods in
  Natural Language Processing}. Association for Computational Linguistics,
  2021.
\newblock URL \url{https://aclanthology.org/2021.emnlp-main.602}.

\bibitem[Lester et~al.(2021)Lester, Al-Rfou, and Constant]{lester2021power}
Brian Lester, Rami Al-Rfou, and Noah Constant.
\newblock The power of scale for parameter-efficient prompt tuning.
\newblock In \emph{Proceedings of the 2021 Conference on Empirical Methods in
  Natural Language Processing (EMNLP)}, 2021.
\newblock URL \url{https://aclanthology.org/2021.emnlp-main.243}.

\bibitem[Lewis et~al.(2021)Lewis, Bhosale, Dettmers, Goyal, and
  Zettlemoyer]{lewis2021base}
Mike Lewis, Shruti Bhosale, Tim Dettmers, Naman Goyal, and Luke Zettlemoyer.
\newblock Base layers: Simplifying training of large, sparse models.
\newblock In \emph{International Conference on Machine Learning}, pages
  6265--6274. PMLR, 2021.

\bibitem[Li and Liang(2021)]{li2021prefix}
Xiang~Lisa Li and Percy Liang.
\newblock Prefix-tuning: Optimizing continuous prompts for generation.
\newblock \emph{arXiv preprint arXiv:2101.00190}, 2021.

\bibitem[Lin et~al.(2020)Lin, Wohlwend, Chen, and
  Lei]{lin-etal-2020-autoregressive}
Alexander Lin, Jeremy Wohlwend, Howard Chen, and Tao Lei.
\newblock Autoregressive knowledge distillation through imitation learning.
\newblock In \emph{Proceedings of the 2020 Conference on Empirical Methods in
  Natural Language Processing (EMNLP)}, 2020.
\newblock URL \url{https://aclanthology.org/2020.emnlp-main.494}.

\bibitem[Martins and Astudillo(2016)]{pmlr-v48-martins16}
Andre Martins and Ramon Astudillo.
\newblock From softmax to sparsemax: A sparse model of attention and
  multi-label classification.
\newblock In \emph{Proceedings of The 33rd International Conference on Machine
  Learning}, 2016.

\bibitem[Mathew et~al.(2021)Mathew, Karatzas, and Jawahar]{mathew2021docvqa}
Minesh Mathew, Dimosthenis Karatzas, and CV~Jawahar.
\newblock Docvqa: A dataset for vqa on document images.
\newblock In \emph{Proceedings of the IEEE/CVF winter conference on
  applications of computer vision}, pages 2200--2209, 2021.

\bibitem[Mishra et~al.(2019)Mishra, Shekhar, Singh, and
  Chakraborty]{mishra2019ocr}
Anand Mishra, Shashank Shekhar, Ajeet~Kumar Singh, and Anirban Chakraborty.
\newblock Ocr-vqa: Visual question answering by reading text in images.
\newblock In \emph{2019 international conference on document analysis and
  recognition (ICDAR)}, pages 947--952. IEEE, 2019.

\bibitem[Narayan et~al.(2018)Narayan, Cohen, and Lapata]{Narayan2018DontGM}
Shashi Narayan, Shay~B. Cohen, and Mirella Lapata.
\newblock Don't give me the details, just the summary! topic-aware
  convolutional neural networks for extreme summarization.
\newblock \emph{ArXiv}, abs/1808.08745, 2018.

\bibitem[Nawrot et~al.(2022)Nawrot, Chorowski, Łańcucki, and
  Ponti]{nawrot2022dynamic}
Piotr Nawrot, Jan Chorowski, Adrian Łańcucki, and Edoardo~M. Ponti.
\newblock Efficient transformers with dynamic token pooling, 2022.

\bibitem[Panayotov et~al.(2015)Panayotov, Chen, Povey, and
  Khudanpur]{panayotov2015librispeech}
Vassil Panayotov, Guoguo Chen, Daniel Povey, and Sanjeev Khudanpur.
\newblock Librispeech: an asr corpus based on public domain audio books.
\newblock In \emph{2015 IEEE international conference on acoustics, speech and
  signal processing (ICASSP)}, pages 5206--5210. IEEE, 2015.

\bibitem[Press et~al.(2019)Press, Smith, and Levy]{press2019improving}
Ofir Press, Noah~A Smith, and Omer Levy.
\newblock Improving transformer models by reordering their sublayers.
\newblock \emph{arXiv preprint arXiv:1911.03864}, 2019.

\bibitem[Raffel et~al.(2020)Raffel, Shazeer, Roberts, Lee, Narang, Matena,
  Zhou, Li, and Liu]{t5}
Colin Raffel, Noam Shazeer, Adam Roberts, Katherine Lee, Sharan Narang, Michael
  Matena, Yanqi Zhou, Wei Li, and Peter~J. Liu.
\newblock Exploring the limits of transfer learning with a unified text-to-text
  transformer.
\newblock \emph{Journal of Machine Learning Research}, 2020.
\newblock URL \url{http://jmlr.org/papers/v21/20-074.html}.

\bibitem[Rajpurkar et~al.(2016)Rajpurkar, Zhang, Lopyrev, and
  Liang]{rajpurkar-etal-2016-squad}
Pranav Rajpurkar, Jian Zhang, Konstantin Lopyrev, and Percy Liang.
\newblock {SQ}u{AD}: 100,000+ questions for machine comprehension of text.
\newblock In \emph{Proceedings of the 2016 Conference on Empirical Methods in
  Natural Language Processing}, 2016.
\newblock URL \url{https://nlp.stanford.edu/pubs/rajpurkar2016squad.pdf}.

\bibitem[Rao et~al.(2021)Rao, Zhao, Liu, Lu, Zhou, and
  Hsieh]{rao2021dynamicvit}
Yongming Rao, Wenliang Zhao, Benlin Liu, Jiwen Lu, Jie Zhou, and Cho-Jui Hsieh.
\newblock Dynamicvit: Efficient vision transformers with dynamic token
  sparsification.
\newblock In \emph{Advances in Neural Information Processing Systems
  (NeurIPS)}, 2021.

\bibitem[Roberts et~al.(2022)Roberts, Chung, Levskaya, Mishra, Bradbury, Andor,
  Narang, Lester, Gaffney, Mohiuddin, Hawthorne, Lewkowycz, Salcianu, van Zee,
  Austin, Goodman, Soares, Hu, Tsvyashchenko, Chowdhery, Bastings, Bulian,
  Garcia, Ni, Chen, Kenealy, Clark, Lee, Garrette, Lee-Thorp, Raffel, Shazeer,
  Ritter, Bosma, Passos, Maitin-Shepard, Fiedel, Omernick, Saeta, Sepassi,
  Spiridonov, Newlan, and Gesmundo]{roberts2022t5x}
Adam Roberts, Hyung~Won Chung, Anselm Levskaya, Gaurav Mishra, James Bradbury,
  Daniel Andor, Sharan Narang, Brian Lester, Colin Gaffney, Afroz Mohiuddin,
  Curtis Hawthorne, Aitor Lewkowycz, Alex Salcianu, Marc van Zee, Jacob Austin,
  Sebastian Goodman, Livio~Baldini Soares, Haitang Hu, Sasha Tsvyashchenko,
  Aakanksha Chowdhery, Jasmijn Bastings, Jannis Bulian, Xavier Garcia, Jianmo
  Ni, Andrew Chen, Kathleen Kenealy, Jonathan~H. Clark, Stephan Lee, Dan
  Garrette, James Lee-Thorp, Colin Raffel, Noam Shazeer, Marvin Ritter, Maarten
  Bosma, Alexandre Passos, Jeremy Maitin-Shepard, Noah Fiedel, Mark Omernick,
  Brennan Saeta, Ryan Sepassi, Alexander Spiridonov, Joshua Newlan, and Andrea
  Gesmundo.
\newblock Scaling up models and data with $\texttt{t5x}$ and $\texttt{seqio}$.
\newblock \emph{arXiv preprint arXiv:2203.17189}, 2022.
\newblock URL \url{https://arxiv.org/abs/2203.17189}.

\bibitem[Roller et~al.(2021)Roller, Sukhbaatar, Weston, et~al.]{roller2021hash}
Stephen Roller, Sainbayar Sukhbaatar, Jason Weston, et~al.
\newblock Hash layers for large sparse models.
\newblock \emph{Advances in Neural Information Processing Systems},
  34:\penalty0 17555--17566, 2021.

\bibitem[Schmitzer(2019)]{schmitzer2019stabilized}
Bernhard Schmitzer.
\newblock Stabilized sparse scaling algorithms for entropy regularized
  transport problems.
\newblock \emph{SIAM Journal on Scientific Computing}, 41\penalty0
  (3):\penalty0 A1443--A1481, 2019.

\bibitem[Schuster et~al.(2022)Schuster, Fisch, Gupta, Dehghani, Bahri, Tran,
  Tay, and Metzler]{schuster2022confident}
Tal Schuster, Adam Fisch, Jai Gupta, Mostafa Dehghani, Dara Bahri, Vinh Tran,
  Yi~Tay, and Donald Metzler.
\newblock Confident adaptive language modeling.
\newblock \emph{Advances in Neural Information Processing Systems}, 2022.

\bibitem[Shazeer(2019)]{shazeer2019fast}
Noam Shazeer.
\newblock Fast transformer decoding: One write-head is all you need.
\newblock \emph{arXiv preprint arXiv:1911.02150}, 2019.

\bibitem[Shazeer(2020)]{shazeer2020glu}
Noam Shazeer.
\newblock Glu variants improve transformer.
\newblock \emph{arXiv preprint arXiv:2002.05202}, 2020.

\bibitem[Shazeer et~al.(2017)Shazeer, Mirhoseini, Maziarz, Davis, Le, Hinton,
  and Dean]{shazeer2017outrageously}
Noam Shazeer, Azalia Mirhoseini, Krzysztof Maziarz, Andy Davis, Quoc Le,
  Geoffrey Hinton, and Jeff Dean.
\newblock Outrageously large neural networks: The sparsely-gated
  mixture-of-experts layer.
\newblock \emph{arXiv preprint arXiv:1701.06538}, 2017.

\bibitem[Shen et~al.(2019)Shen, Nguyen, Wu, Chen, et~al.]{shen2019lingvo}
Jonathan Shen, Patrick Nguyen, Yonghui Wu, Zhifeng Chen, et~al.
\newblock Lingvo: a modular and scalable framework for sequence-to-sequence
  modeling, 2019.

\bibitem[So et~al.(2021)So, Ma{\'n}ke, Liu, Dai, Shazeer, and
  Le]{so2021searching}
David So, Wojciech Ma{\'n}ke, Hanxiao Liu, Zihang Dai, Noam Shazeer, and Quoc~V
  Le.
\newblock Searching for efficient transformers for language modeling.
\newblock \emph{Advances in Neural Information Processing Systems},
  34:\penalty0 6010--6022, 2021.

\bibitem[Su et~al.(2021)Su, You, Xie, Zheng, Wang, Qian, Zhang, Wang, and
  Xu]{su2021vision}
Xiu Su, Shan You, Jiyang Xie, Mingkai Zheng, Fei Wang, Chen Qian, Changshui
  Zhang, Xiaogang Wang, and Chang Xu.
\newblock Vision transformer architecture search.
\newblock \emph{arXiv e-prints}, pages arXiv--2106, 2021.

\bibitem[Tai et~al.(2022)Tai, Tian, and Lim]{tai2022spartan}
Kai~Sheng Tai, Taipeng Tian, and Ser-Nam Lim.
\newblock {Spartan: Differentiable Sparsity via Regularized Transportation}.
\newblock In \emph{Advances in Neural Information Processing Systems}, 2022.

\bibitem[Turc et~al.(2019)Turc, Chang, Lee, and Toutanova]{turc2019well}
Iulia Turc, Ming-Wei Chang, Kenton Lee, and Kristina Toutanova.
\newblock Well-read students learn better: On the importance of pre-training
  compact models.
\newblock \emph{arXiv preprint arXiv:1908.08962}, 2019.

\bibitem[Vaswani et~al.(2017)Vaswani, Shazeer, Parmar, Uszkoreit, Jones, Gomez,
  Kaiser, and Polosukhin]{vaswani2017attention}
Ashish Vaswani, Noam Shazeer, Niki Parmar, Jakob Uszkoreit, Llion Jones,
  Aidan~N Gomez, {\L}ukasz Kaiser, and Illia Polosukhin.
\newblock Attention is all you need.
\newblock \emph{Advances in neural information processing systems}, 30, 2017.

\bibitem[Vu et~al.(2022)Vu, Barua, Lester, Cer, Iyyer, and
  Constant]{vu2022overcoming}
Tu~Vu, Aditya Barua, Brian Lester, Daniel Cer, Mohit Iyyer, and Noah Constant.
\newblock Overcoming catastrophic forgetting in zero-shot cross-lingual
  generation.
\newblock \emph{arXiv preprint arXiv:2205.12647}, 2022.

\bibitem[Wang et~al.(2021)Wang, Li, Zhou, Chen, Grossman, and Li]{screen2words}
Bryan Wang, Gang Li, Xin Zhou, Zhourong Chen, Tovi Grossman, and Yang Li.
\newblock Screen2words: Automatic mobile ui summarization with multimodal
  learning.
\newblock In \emph{The 34th Annual ACM Symposium on User Interface Software and
  Technology}, UIST '21, page 498–510, New York, NY, USA, 2021. Association
  for Computing Machinery.
\newblock ISBN 9781450386357.
\newblock \doi{10.1145/3472749.3474765}.
\newblock URL \url{https://doi.org/10.1145/3472749.3474765}.

\bibitem[Wang et~al.(2020{\natexlab{a}})Wang, Li, Khabsa, Fang, and
  Ma]{wang2020linformer}
Sinong Wang, Belinda~Z Li, Madian Khabsa, Han Fang, and Hao Ma.
\newblock Linformer: Self-attention with linear complexity.
\newblock \emph{arXiv preprint arXiv:2006.04768}, 2020{\natexlab{a}}.

\bibitem[Wang et~al.(2020{\natexlab{b}})Wang, Wohlwend, and
  Lei]{wang-etal-2020-structured}
Ziheng Wang, Jeremy Wohlwend, and Tao Lei.
\newblock Structured pruning of large language models.
\newblock In \emph{Proceedings of the 2020 Conference on Empirical Methods in
  Natural Language Processing (EMNLP)}. Association for Computational
  Linguistics, 2020{\natexlab{b}}.
\newblock URL \url{https://aclanthology.org/2020.emnlp-main.496}.

\bibitem[Williams et~al.(2018)Williams, Nangia, and
  Bowman]{williams-etal-2018-broad}
Adina Williams, Nikita Nangia, and Samuel Bowman.
\newblock A broad-coverage challenge corpus for sentence understanding through
  inference.
\newblock In \emph{Proceedings of the 2018 Conference of the North {A}merican
  Chapter of the Association for Computational Linguistics: Human Language
  Technologies, Volume 1 (Long Papers)}, 2018.
\newblock URL \url{https://aclanthology.org/N18-1101}.

\bibitem[Wright(2015)]{wright2015coordinate}
Stephen~J Wright.
\newblock Coordinate descent algorithms.
\newblock \emph{Mathematical Programming}, 151\penalty0 (1):\penalty0 3--34,
  2015.

\bibitem[Xia et~al.(2022)Xia, Zhong, and Chen]{xia2022structured}
Mengzhou Xia, Zexuan Zhong, and Danqi Chen.
\newblock Structured pruning learns compact and accurate models.
\newblock In \emph{Association for Computational Linguistics (ACL)}, 2022.

\bibitem[Xie et~al.(2020)Xie, Luong, Hovy, and Le]{xie2020self}
Qizhe Xie, Minh-Thang Luong, Eduard Hovy, and Quoc~V Le.
\newblock Self-training with noisy student improves imagenet classification.
\newblock In \emph{Proceedings of the IEEE/CVF conference on computer vision
  and pattern recognition}, pages 10687--10698, 2020.

\bibitem[Yin et~al.(2022)Yin, Vahdat, Alvarez, Mallya, Kautz, and
  Molchanov]{Yin_2022_CVPR}
Hongxu Yin, Arash Vahdat, Jose~M. Alvarez, Arun Mallya, Jan Kautz, and Pavlo
  Molchanov.
\newblock {A}-{V}i{T}: {A}daptive tokens for efficient vision transformer.
\newblock In \emph{Proceedings of the IEEE/CVF Conference on Computer Vision
  and Pattern Recognition (CVPR)}, pages 10809--10818, June 2022.

\bibitem[Zhang et~al.(2022)Zhang, Park, Han, Qin, Gulati, Shor, Jansen, Xu,
  Huang, Wang, et~al.]{zhang2022bigssl}
Yu~Zhang, Daniel~S Park, Wei Han, James Qin, Anmol Gulati, Joel Shor, Aren
  Jansen, Yuanzhong Xu, Yanping Huang, Shibo Wang, et~al.
\newblock Bigssl: Exploring the frontier of large-scale semi-supervised
  learning for automatic speech recognition.
\newblock \emph{IEEE Journal of Selected Topics in Signal Processing},
  16\penalty0 (6):\penalty0 1519--1532, 2022.

\bibitem[Zhou et~al.(2022)Zhou, Lei, Liu, Du, Huang, Zhao, Dai, Chen, Le, and
  Laudon]{expertchoice2022}
Yanqi Zhou, Tao Lei, Hanxiao Liu, Nan Du, Yanping Huang, Vincent Zhao, Andrew
  Dai, Zhifeng Chen, Quoc Le, and James Laudon.
\newblock Mixture-of-experts with expert choice routing, 2022.
\newblock URL \url{https://arxiv.org/abs/2202.09368}.

\bibitem[Zhu and Gupta(2017)]{zhu2017prune}
Michael Zhu and Suyog Gupta.
\newblock To prune, or not to prune: exploring the efficacy of pruning for
  model compression.
\newblock \emph{arXiv preprint arXiv:1710.01878}, 2017.

\end{thebibliography}
